%% file: elsarticle-template.tex
\documentclass[review]{elsarticle}

\usepackage{graphicx}

\graphicspath{{Figures/}{Figures/Approach/}{Figures/Chapter4/qualitative_results/iseg/}{Figures/Chapter4/qualitative_results/BSD100/}{Figures/Chapter4/qualitative_results/horses/}{Figures/Chapter4/qualitative_results/wzm1/}{Figures/Chapter4/qualitative_results/wzm2/}{Figures/Chapter4/robot_user/}
{Figures/why_it_works/}
{Figures/Introduction/}}
\DeclareGraphicsExtensions{.pdf,.jpeg,.jpg,.eps}

\usepackage{comment}
\usepackage{tabularx}
\usepackage{amsmath}
\usepackage{multirow}
\usepackage{dblfloatfix}
\usepackage{algorithmicx}
\usepackage{algpseudocode}
\usepackage{soul}
\usepackage[ruled]{algorithm}
\usepackage{algpseudocode}
\usepackage{pgfplots}
\usepackage{amssymb}
\usepgfplotslibrary{fillbetween}
\usepackage{lineno,hyperref}
\usepackage[caption=false,font=footnotesize]{subfig}
\modulolinenumbers[5]

\begin{document}

\begin{frontmatter}

\title{Seeded Laplaican: An Eigenfunction Solution for Scribble Based Interactive Image Segmentation}


\author[umd]{Ahmed Taha\corref{cor1}}
\ead{ahmdtaha@cs.umd.edu}

\author[alex]{Marwan Torki}
\ead{mtorki@alexu.edu.eg}

\cortext[cor1]{Corresponding author}

\address[umd]{University of Maryland, College Park}
\address[alex]{Alexandria University, Faculty of Engineering}




\input{abstract}


\end{frontmatter}


\ifnum 2>1
\input{Introduction_review}
\input{Related_Work_review}

\input{Semi_Supervised_Learning_review}

\input{Approach_review}
\input{experimental_results_review}
\else
\input{Introduction}
\input{Related_Work}
\input{Semi_Supervised_Learning}
\input{Approach}
\input{experimental_results}
\fi

\section{Conclusion}
We presented Seeded Laplacian (SL), a scribble-based interactive image segmentation approach. The image segmentation problem is cast as a graph-based, semi-supervised learning problem. We optimized the laplacian eigenfunction computation procedure to fit the time-constrained nature of the problem. To generalize our approach, we created five newly annotated scribbled-based datasets.  We studied different pixel features to identify the best for image segmentation purposes. The features proposed are the cornerstone of SL superiority. Spatial features with geodesic distance integration appeared noteworthy in our experiments. SL can be easily extended with other feature vectors like depth and texture. The experimental results section provides quantitative and qualitative evidence of SL effectiveness against state-of-the-art algorithms. Robot user analysis shows SL ability to adapt with a flexible sequence of user interactions in a precise manner.


\section*{References}
\bibliography{refs}

\end{document}

%% file: abstract.tex
\begin{abstract}

In this paper, we cast the scribble-based interactive image segmentation as a semi-supervised learning problem.  Our novel approach alleviates the need to solve an expensive generalized eigenvector problem  by approximating the eigenvectors solution using efficiently computed eigenfunctions. The smoothness operator defined on feature densities at the limit $n\rightarrow\infty$ recovers the exact eigenvectors of the graph Laplacian, where $n$ is the number of nodes in the graph. To further reduce the computational complexity without scarifying our accuracy, we select pivot pixels from user annotations. Through control experiments, prime segmentation features combination is proposed. This combination is the key pillar for our approach superiority and reliability across five datasets.


In our experiments, we evaluate our approach using both human scribble and  ``robot user'' annotations to guide the foreground/background segmentation. We developed new unbiased collection of five annotated images datasets to standardize the evaluation procedure for scribble-based segmentation methods. We experimented with several variations, including different feature vectors, pivots count and the number of eigenvectors. Experiments are carried out on datasets that contain a wide variety of natural images. We achieve better qualitative and quantitative results compared to state-of-the-art interactive segmentation algorithms. 

\end{abstract}


%% file: Introduction_review.tex
\section{Introduction}\label{sec:introduction}

%
%
%
%

\label{sec:intro}

Image segmentation is an important problem in computer vision. It is a common intermediate step in image processing; image segmentation divides an image into a small set of meaningful segments that simplify further analysis.  More precisely, image segmentation is the process of grouping pixels sharing certain visual characteristics into separate regions.  Some of the practical applications of image segmentation are processing medical images ~\cite{pham2000current,grau2004improved} and satellite images ~\cite{bins1996satellite,pesaresi2001new} to locate objects. Content-based image retrieval ~\cite{rui1999image,belongie1998color} is another important application for image segmentation algorithms.

In this paper, we model interactive image segmentation as a graph-based semi-supervised learning problem. We calculate Laplace-Beltrami eigenfunctions to approximate Laplacian eigenvectors. Such trick reduces the space and time needed to build and solve the graph-based labeling process considerably, from minutes to seconds. To the best of our knowledge, we are the first to solve the scribble-based interactive segmentation problem using efficiently computed eigenfunctions.

\begin{figure}
\subfloat{\includegraphics[width=0.2\textwidth,height=0.2\textwidth]{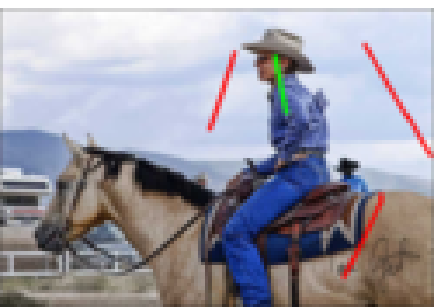}}   \hfill
	\subfloat{\includegraphics[width=0.2\textwidth,height=0.2\textwidth]{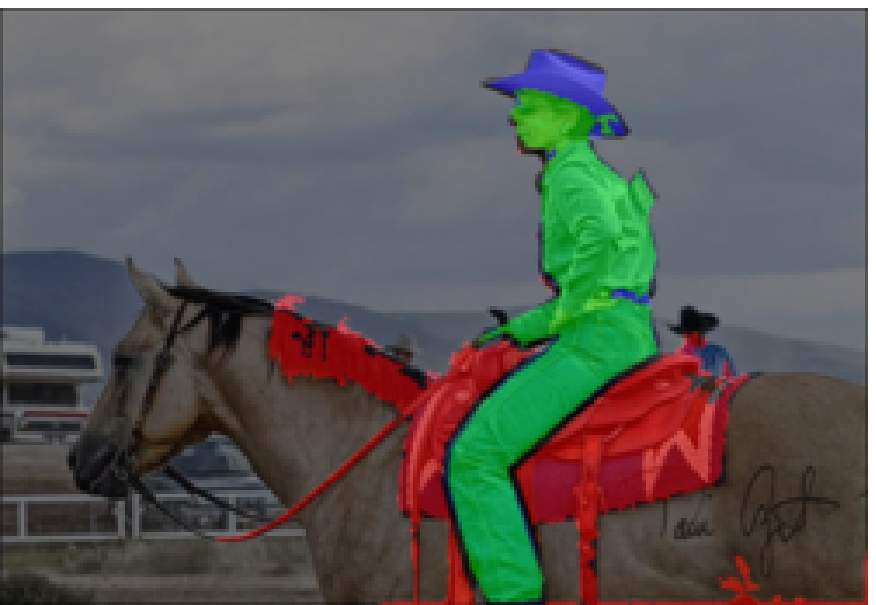}}   \hfill
	\subfloat{\includegraphics[width=0.2\textwidth,height=0.2\textwidth]{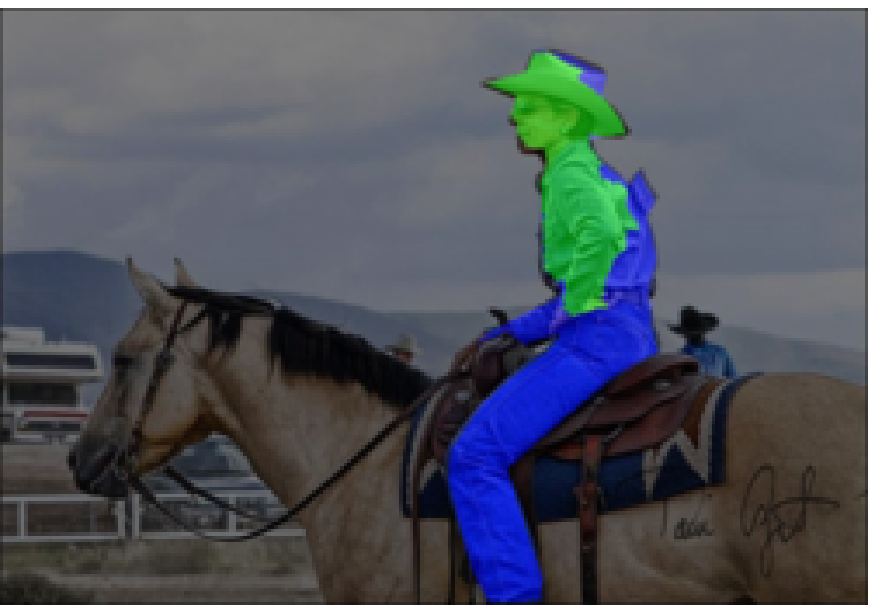}}   \hfill
	\subfloat{\includegraphics[width=0.2\textwidth,height=0.2\textwidth]{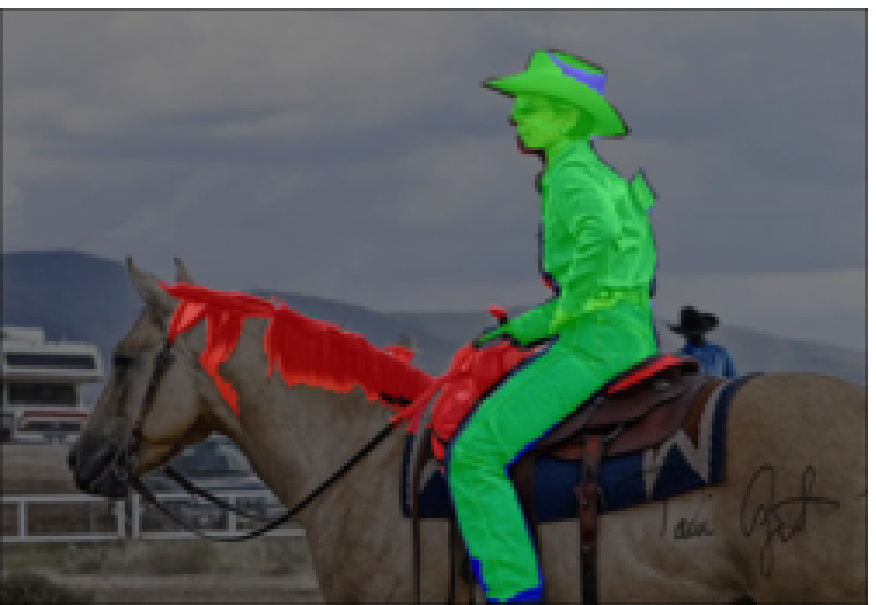}}  	
	
	\subfloat[User Scribble]{\includegraphics[width=0.2\textwidth, height=0.25\textwidth]{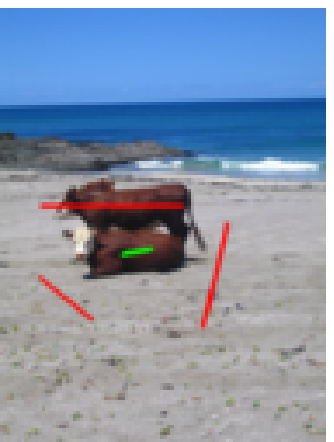}}   \hfill
	\subfloat[RGB]{\includegraphics[width=0.2\textwidth, height=0.25\textwidth]{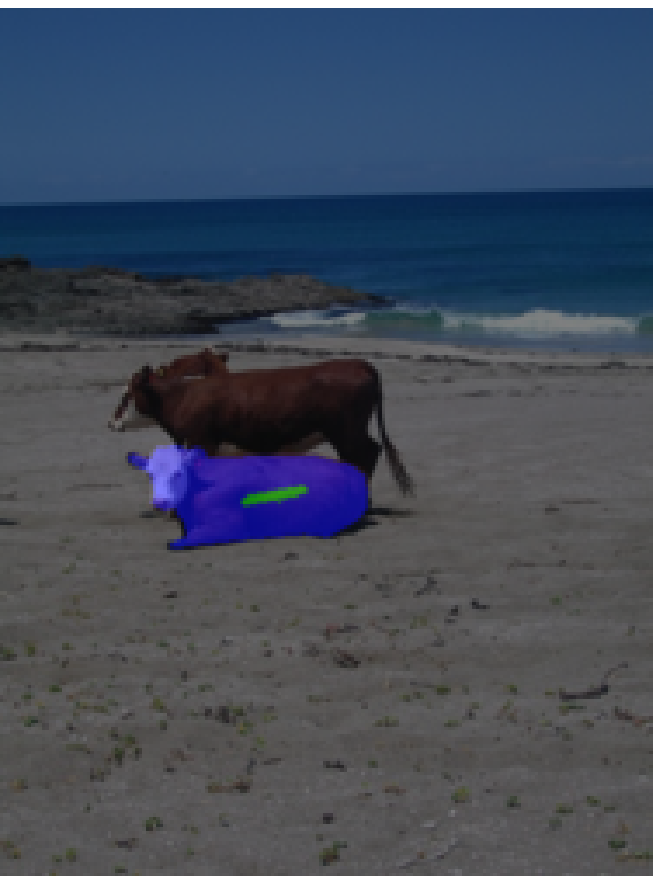}}   \hfill
	\subfloat[Geodesic]{\includegraphics[width=0.2\textwidth, height=0.25\textwidth]{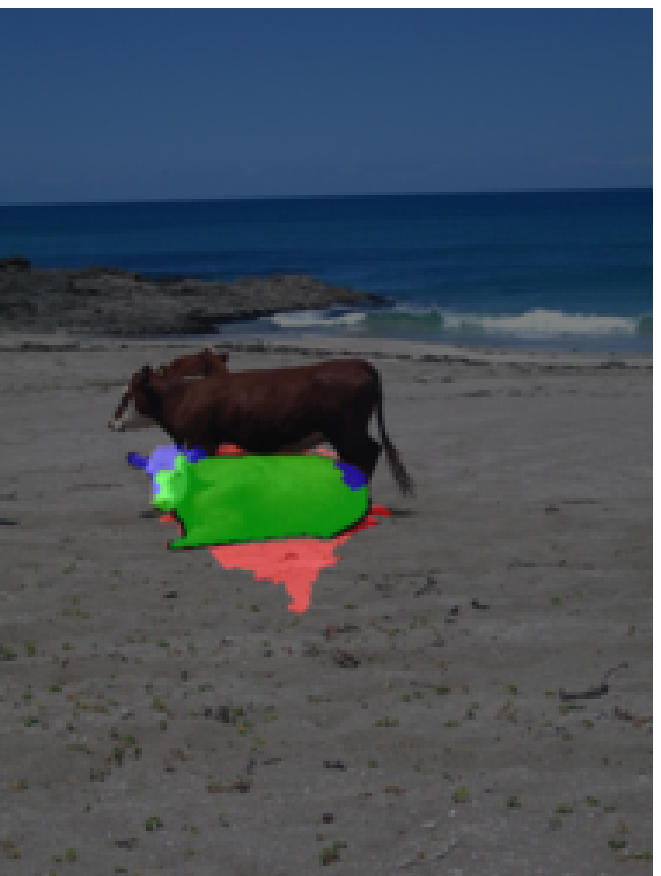}}    \hfill
	\subfloat[Our Features]{\includegraphics[width=0.2\textwidth, height=0.25\textwidth]{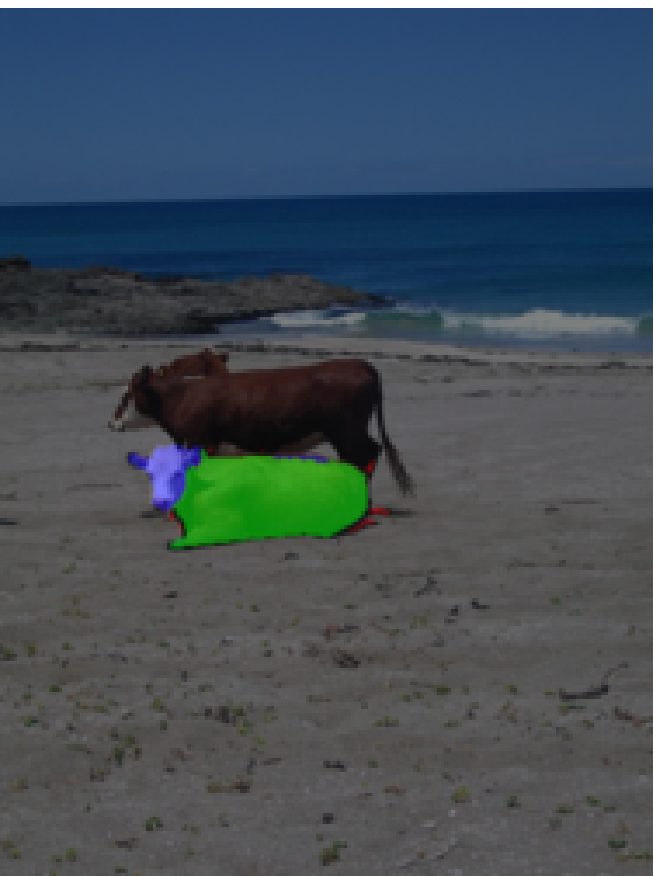}}   
	
	\caption[Segmentation Feature Comparison]{Different features used for segmentation. First column shows user scribbles. The other columns show the segmentation result using naive color (RGB), geodesic and our features combination. Green, red and blue overlays indicate true positive, false positive and false negative respectively.}\label{fig:seg_ft_extraction}	
\end{figure}

We present a \textit{novel} scribble-based interactive image segmentation algorithm, Seeded Laplacian (SL), for foreground/background segmentation. Unlike image labeling problem ~\cite{fergus2009semi}, interactive segmentation is both a latency and accuracy constrained. Our approach builds upon Fergus et al.~\cite{fergus2009semi} work in two different directions.  We tackle the latency constraint using a two-fold technique to speedup the segmentation process. This is achieved by matrix operation optimization and user scribble sampling. We tackle the accuracy constraint by proposing better image segmentation features. Figure ~\ref{fig:seg_ft_extraction} shows how naive RGB color or geodesic features can lead to hazy segmentation results. So we investigated different features combinations that generalize well on natural images.

Thus, our formulation outperforms well-known segmentation approaches on five different datasets and  brings three \textbf{\textit{key contributions}} to the problem: \\
\textbf{1) Scalability:}  The exact eigenvectors computation of a graph Laplacian is space and time consuming. Despite being faster to compute, the eigenfunctions are still slow for interactive problems.  So we propose an optimized eigenfunction computation version and scribble sampling technique. This drastically reduces the time needed to achieve real-time performance.

\noindent\textbf{2) Accuracy:} SL achieves highly competitive results against state-of-the-art interactive image segmentation methods. We also release \textbf{a collection of five newly annotated datasets} to generalize our SL approach. 


\noindent\textbf{3) Flexibility to feature type:} SL supports different pixel features like spatial information, different color spaces, geodesic distance, and intervening contour. It can be extended easily to incorporate new features like depth, texture or patch descriptors.\\

Unlike unsupervised learning, we guide our learning problem with user provided scribbles. Semi-supervised learning is also considered to be more adequate than supervised learning for scribble image problem. While supervised learning approaches depends on a large labeled set to generate a highly accurate prediction,  semi-supervised learning can benefit from a small set of labeled data like user scribbles.
  
Semi-supervised learning can also benefit from the distribution of the labeled and unlabeled pixels. Thus, similarity measures between unlabeled pixels contribute to our learning problems unlike the supervised learning approach.

The interactive image segmentation problem holds all the assumptions required by semi-supervised learning ~\cite{chapelle2006semi}.

\textit{1.\underline{Smoothness assumption}}: If two points $x_1, x_2$ in a high-density region are close, then the corresponding labeling $y_1, y_2$ should also be close. Such an assumption is valid for image pixels because foreground and background pixels lie close to each other in high density regions in feature space.

\textit{2.\underline{Low density separation (a.k.a Cluster assumption)}}: The decision boundary should lie in a low-density region. In the foreground image segmentation problem, the foreground object is separated from the background through a boundary contour lying in a low-density region.

\begin{figure}
	\centering
	{\includegraphics[width=1\textwidth]{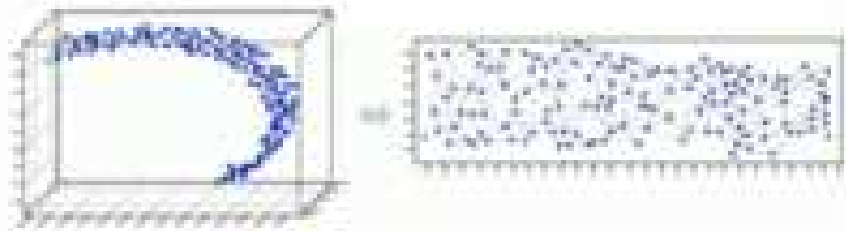}}    
	\caption{Three Dimensional mainfold embedding into two dimensional graph.}\label{fig:mainfold_assumption}
\end{figure}

\textit{3.\underline{Manifold assumption}}: The high-dimensional data can be mapped on a low-dimensional manifold as shown in figure ~\ref{fig:mainfold_assumption}. In our approach, the pixels of the image are embedded in 2D Laplacian graph matrix. Such graph matrix encapsulates the relationship between image pixels.

Such homogeneity between the semi-supervised learning assumptions and the interactive image segmentation problem supports our cast.


%% file: Related_Work_review.tex
\section{Related Work}

Due to the difficulty of fully automatic image segmentation, user-interactive segmentation is usually introduced to relax the segmentation problem for certain applications. In interactive image segmentation, users guide the segmentation process by providing annotations. User-specific annotations can take various forms, e.g., bounding box ~\cite{rother2004grabcut}, sloppy contour ~\cite{mortensen1995intelligent,kass1988snakes}, and scribbles ~\cite{boykov2001interactive}. Although, scribbles are often favored due to their ease of use in terms of time and effort, scribbles generally provide less information than  bounding box or sloppy contour. There is always a compromise between the choice of  annotation type in terms of speed and its effect on the quality and accuracy of the segmentation process. A recent study ~\cite{jain2013predicting} predicts the easiest input annotation form that will be sufficiently strong to successfully segment a given image.

In the following we describe and compare several well-known interactive scribble segmentation methods. Scribble segmentation methods can be categorized into two main categories: \textbf{region growing-based} methods and \textbf{graph-based} methods. In region growing methods, an iterative approach is employed to label unlabeled pixels near the labeled ones. This iterative process ends when all pixels are labeled as either foreground or background pixels. Known examples for the region growing methods include Maximal Similarity-based Region Merging (MSRM) ~\cite{ning2010interactive} and seeded region growing ~\cite{adams1994seeded}.
On the other hand, graph-based methods like normalized cuts \cite{shi2000normalized} and Boykov Jolly ~\cite{boykov2001interactive} have clear cost function; but they are computationally expensive. Fortunately, fast implementations of polynomial graph cut algorithms are available.

MSRM ~\cite{ning2010interactive} is a well-known region growing-based method. It requires an initial partitioning of an image into homogeneous regions. Given initial segmentation (super-pixels), usually using the mean-shift method ~\cite{cheng1995mean}, MSRM calculates a color histogram for each super-pixel. Using the user seeded background and foreground annotations, regions are categorized into background, foreground, or unknown regions. MSRM iterates over the unknown regions and calculates the Bhattacharyya coefficient ~\cite{kailath1967divergence} $\rho (Q,R)$ to measure the similarity between two regions, R and Q. Based on the Bhattacharyya coefficient $\rho (Q,R)$, unknown regions are either marked as foreground or background accordingly.

Region growing methods encounter a number of drawbacks. For example, they do not have a clear cost function. They also suffer when the foreground or background regions are not connected regions and require extra user annotation to overcome this limitation. Being iterative is yet another computational limitation for these methods, but using super-pixels is a typical workaround for this obstacle.

On the other hand, graph-cut based methods have a clear cost function; they do not suffer from the unconnected regions problem but they are computationally expensive. Fortunately, fast implementations of polynomial graph cut algorithms are available, like max-flow ~\cite{ford1962flows}, push-relabel ~\cite{goldberg1988new} and eigenvector approximation for graph Laplacian \cite{shi2000normalized}.
 
Normalized cuts \cite{shi2000normalized} is one of those graph-based methods, it aims to partition the graph $V$ into two partitions $A$ and $B$ such that the graph cut cost is as minimal as possible.

\begin{equation} \label{eq:graph-cut-cost}
\text{cut cost}=\frac { cut(A,B) }{ assoc(A,V) } +\frac { cut(A,B) }{ assoc(B,V) } 
\end{equation}
Where $cut(A, B)$ is the sum of weights of all edges that has one end in $A$ and another end in $B$, and $assoc(A,V)$ is the sum of weights of all edges that has one end in $A$. The cost of cut is small when the weight of edges connecting $A$ and $B$ is very small, while the weights of edges inside $A$ and $B$ are big.

Solving eq. ~\ref{eq:graph-cut-cost} is computationally very expensive and sometimes not feasible, so an approximate solution was introduced by solving the generalized eigenvector  eq. ~\ref{eq:normalised-cuts} to generate an approximate graph cut.

\begin{equation} \label{eq:normalised-cuts}
(D - W)v =\lambda Dv  
 \end{equation}
Where $W$ is the Affinity Matrix and $D$ is the Degree Matrix
 \begin{equation}
\begin{split}
 { W }_{ i,j }={ e }^{ -\frac { \left\| { x }_{ i }-{ x }_{ j } \right\|  }{ t }  } \\
{ D }_{ i,i } =\sum _{ j=1 }^{ n }{ {W}_{i,j} } ,i\neq j 
\end{split}
 \end{equation}

Boykov-Jolly ~\cite{boykov2001interactive} is another graph cut based method. By constructing a graph in a fashion similar to the normalized cuts method, ~\cite{boykov2001interactive} Boykov-Jolly tries to minimize the cost function $E(A)$.

\begin{equation} \label{eq:boykov-jolly}
E(A) =\lambda . R(A) + B(A)  
 \end{equation}
Where \begin{equation}
 R(A)=\sum_{p\in P} { { R }_{ p\in P } } \quad , \quad  B(A)=\sum _{ \{ p,q\in N\}  } { { B }_{ \{ p,q\}  } . \delta(A_p,A_q) } 
 \end{equation}
and 
\begin{equation}
{ \delta (A_{ p },A_{ q }) }=\begin{cases} 1 & if\quad A_{ p }\neq A_{ q } \\ 0 & otherwise \end{cases}
 \end{equation}

Where $A = (A_1,A_2,......,A_{|p|})$ is a binary vector whose components $A_p$ specify assignments to pixels $p$ in $P$. Each $A_p$ can be belong to ``Object'' or ``Background''. The coefficient $\lambda \ge  0$ specifies a relative importance of the region properties term  $R(A)$ versus the boundary properties term $B(A)$. The regional term $R(A)$ assumes that the individual penalties for assigning pixel $p$ to ``object" and ``background", correspondingly $R_p(``obj")$ $R_p(``pkg")$, are given. For example, $R_p(.)$ may reflect how the intensity of pixel $p$ fits into a known intensity model (e.g. histogram) of the object and background.

\begin{equation}
{ B }_{ \{ p,q\}  }\quad \propto \quad exp\left( -\frac { { (I_{ q }\quad -\quad I_{ p }) }^{ 2 } }{ 2{ \sigma  }^{ 2 } }  \right) .\frac { 1 }{ dist(p,q) }.
 \end{equation}

The term $B(A)$ comprises the “boundary” properties of segmentation $A$. Coefficient $B_{\{p,q\}} \ge 0$ should be interpreted as a penalty for a discontinuity between pixels $p$ and $q$. Normally, $B_{\{p,q\}}$ is large when pixels $p$ and $q$ are similar and $B_{\{p,q\}}$ is close to zero when the two are very different. The penalty $B_{\{p,q\}}$ can also decrease as a function of distance between $p$ and $q$.

In ~\cite{gulshan2010geodesic}, Gulshan et al. proposed a shape-constrained graph-based segmentation algorithm. They combined star-convexity constraints with the graph cut energy equation formulated by Boykov-Jolly ~\cite{boykov2001experimental}. Thus, the global minima of the energy equation is subject to the star-convexity constraints. They extended Veksler's work ~\cite{veksler2008star} in two directions: 1) single star convexity was extended to multiple star convexity support, and 2) a geodesic path was suggested as an alternative for Euclidean rays. Gulshan et al. used the user scribbles as the shape star centers, and a sequential system was developed so the shape constraints change progressively with user interaction.

\textbf{Color Features}: Many algorithms in the literature try to solve the Fg/Bg segmentation problem based on the color features. From the region growing family, the seeded region growing algorithm~\cite{adams1994seeded} iterates to assign a pixel to its nearest labeled point based on color distance. In MSRM~\cite{ning2010interactive}, color histograms are built on top of pre-computed super-pixels, and the unlabeled regions are merged to similarly labeled regions using the Bhattacharyya coefficient as a similarity measure. 
Color features are also utilized in the graph cut family. For example, in  grab cut~\cite{rother2004grabcut}, multiple color Gaussian Mixture Models are introduced for each foreground and  background. These color models are found using an iterative procedure that alternates between estimation and parameter learning.\\
\textbf{Geodesic Distance}: Using geodesic distance proved to be useful in interactive image segmentation. Many approaches, like ~\cite{bai2007geodesic, criminisi2008geos, gulshan2010geodesic}, used the definition of the geodesic distance $GeoDist(x)$ as the smallest integral of a weight function over all paths from the scribbles to pixel $x$. These methods compute the $GeoDist(x)$ per class, where $I\in \{ Fg,Bg\}$. There are fast algorithms ~\cite{toivanen1996new,yatziv2006n} that compute the $GeoDist(x)$ in $O(N)$, where N is the number of pixels. Following the same notation as ~\cite{gulshan2010geodesic,criminisi2008geos}, we define length of a discrete path as:
\begin{equation}
L(\Gamma )= \sum _{ i=1 }^{ n-1 }{ \sqrt { (1-{ \gamma }_{ g }){ d({ \Gamma }^{ i },{ \Gamma }^{ i+1 }) }^{ 2 } + { \gamma }_{ g }{ \parallel \triangledown I({ \Gamma }^{ i })\parallel }^{ 2 } } }
\end{equation}
where $\Gamma$ is an arbitrary parametrized discrete path with n pixels given by $\left\{ {\Gamma}^{1},{\Gamma}^{2},.....,{\Gamma}^{n} \right\}$. $d({\Gamma}^{i},{\Gamma}^{i+1})$ is the Euclidean distance between successive pixels, and the quantity ${\parallel \triangledown I({ \Gamma }^{ i })\parallel}^{2} $ is a finite difference approximation of the image gradient between the points $({\Gamma}^{i},{\Gamma}^{i+1})$. The parameter $ { \gamma }_{ g }$ weights the Euclidean distance with the geodesic length.

\begin{figure}[h!]
	\centering\subfloat[Canny Edge result]{\includegraphics[width=0.25\textwidth]{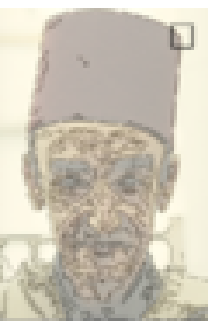}}    ~
	\centering\subfloat[Zoom in rectangle]{\includegraphics[width=0.25\textwidth]{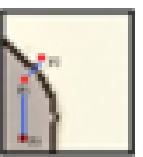}}    
	
	\caption[Intervening Contour]{Intervening contour demonstration. First we compute edges through Canny edge detector. Then we measure the affinity between pixles based on the intervening contours separating them. Affinity between $P_1$ and $P_3$ is higher than $P_1$ and $P_2$  due to the existence of intervening contour between $P_1$ and $P_2$.}\label{fig:ic}
\end{figure}

Using the above definition, one can define the geodesic distance as
\begin{equation}
{d}_{ g }(a,b)=\min _{ \Gamma \in { P}_{ a,b } }{ L(\Gamma ) } , { \Gamma }_{ a,b }= arg \min _{ \Gamma \in { P }_{ a,b } }{ L(\Gamma ) }
\end{equation}
Where ${ P }_{ a,b }$ is the set of all paths between pixels $a, b$. A Path $P$ is defined as a sequence of spatially neighbouring points in 8-connectivity. Distance between neighbouring pixels takes into consideration the spatial distance and color difference. Thus, if the distance between pixels a, b is small, there is a path between $a, b$ along which the color varies only slightly.
\\
\textbf{Intervening contour}: Previous work ~\cite{taha2015seeded} proposed intervening contour as a well-suited interactive segmentation feature vector. It provides better segmentation at object's boundaries. The intuition behind intervening contour is that if the pixels lie in different segments, then we expect to find an intervening contour somewhere along the line ~\cite{malik2001contour}. If no such discontinuity is encountered, then the affinity between the pixels should be large.  Following ~\cite{malik2001contour}, the intervening contour is defined as follows:
\begin{equation}{ W }_{ ij }^{ IC }\quad=\max _{ x\in { M }_{ ij } }{ { p }_{ con }(x) } \end{equation}
where $M_{ij}$ is the set of local maxima along the line joining pixels $i$ and $j$ and $ 0 <  {p }_{ con }(x) < 1$. In order to compute the intervening contour cue, we require a boundary detector that works robustly on natural images. For this we employ the Canny gradient-based boundary detector ~\cite{canny1986computational}.\\
Figure ~\ref{fig:ic} demonstrates the intervening contour intuition. First Canny edge is computed, then affinity between pixels is calculated based on separating contours. 

%% file: Semi_Supervised_Learning_review.tex
\section{Semi-Supervised Learning}
In our approach, we model the interactive image segmentation problem as a semi-supervised learning problem.  Following the notations of Zhu et al ~\cite{zhu2003semi}, the user provides  labeled points, pixels in our case, of input-output pairs $(X_l,Y_l) = \{(x_1,y_1),...,(x_l,y_l)\} $ and unlabeled pixels $X_u = \{x_{l+1}, ..., x_n\}$. In our problem, \begin{math}Y_l\in\{B, F\}\end{math}, where $B$ denotes a background label and $F$ denotes a foreground label.


A very common approach in semi-supervised learning is to use a graph-based algorithm. In graph-based methods , a graph $G = (V,E)$ is constructed where the vertices $V$ are the pixels $x_1 , ..., x_n$ , and the edges $E$ are represented by an $n \times n$ matrix $W$ . Entry $W_{ij}$ is the edge weight between pixels $x_i,x_j$ and a common practice is to set $W_{ij} =exp( { -\left\| { x }_{ i }-{ x }_{ j } \right\|  }^{ 2 }/2{ \epsilon  }^{ 2 } ) $. Let $D$ be a diagonal matrix whose diagonal elements are given by $D_{ii} = \sum _{ j }{W_{ij}}$ , the combinatorial graph Laplacian is defined as $L = D - W$ , which is also called the un-normalized Laplacian. 
A common objective function will have the following form: 
\begin{align} \label{eq:solve}
J(f)&=f^{ T }Lf+\sum _{ i=1 }^{ l }{ \lambda { (f(i)-{ y }_{ i }) }^{ 2 } }  \\
&= f^{ T }Lf +{ (f-y) }^{ T }\Lambda (f-y)
\end{align}
The first term in eq.~\ref{eq:solve} controls the smoothness of the labeling process. This ensures the estimated labels $f_i's$ will not change too much for nearby features in the feature space. The second term penalizes the disagreement between the estimated labels $f_i's$ and the original labels $y_i's$ that are given to the algorithm.

$\Lambda$ is a diagonal matrix whose diagonal elements $\Lambda_{ii}$ equals $\lambda$ if $i$ is a labeled pixel and $\Lambda_{ii} =0$ for unlabeled pixels. The minimizer of eq.~\ref{eq:solve} is the  solution of $ (L + \Lambda)f = \Lambda y.$ To reduce the complexity of the problem, a small number of eigenvectors with the smallest eigenvalues are chosen as suggested by ~\cite{scholkopf2002learning,zhu2003semi,chapelle2006semi}.

As noted by ~\cite{fergus2009semi}, we can significantly reduce the dimension of $f$ by requiring it to be of the form 
\begin{equation} \label{eq:laplacian_smoothness_equation}
f = U\alpha 
\end{equation}
 where $U$ is a $n \times k$ matrix whose columns are the $k$ eigenvectors with smallest eigenvalues. We now have:
\begin{equation}
J(\alpha )={ \alpha  }^{ T }\Sigma \alpha +{ (U\alpha -y) }^{ T }\Lambda (U\alpha -y)
\end{equation}
Where $\Sigma = { U }^{ T }LU$. It can be shown that the minimizing $\alpha$ is now a solution to the $k \times k$ system of equations:
\begin{equation}
(\Sigma +{ U }^{ T }\Lambda U)\alpha ={ U }^{ T }\Lambda y
\end{equation}
\begin{equation}\label{eq:eigvec_solution}
\alpha ={ \left( \Sigma +{ U }^{ T }\Lambda  U\right)  }^{ -1 } ({ U }^{ T }\Lambda y)
\end{equation}
In case of image segmentation, the eigenvector solution is costly. A tiny image of size $100\times 100$ produces an L matrix of size $10000 \times 10000$. Hence the solution for eigenvectors is costly in terms of both space and time. 
\subsection{Eigenfunction Approach}
Like ~\cite{nadler2006diffusion,weiss2009spectral}, we assume $x_i's\in\Re^d$ are samples from a distribution $p(x)$. This density defines a weighted smoothness operator on any function $F(x)$ defined on $\Re^d$, which we denote by: \\$L_p(F)= \frac{1}{2}\int(F(x_1) - F(x_2))^2W(x_1, x_2)p(x_1)p(x_2)dx_1x_2$
Where $W(x_1, x_2) = \exp(−\left\|x_1 - x_2\right\|^2 /2\epsilon^2)$. 

According to~\cite{fergus2009semi}, under suitable convergence conditions the eigenfunctions of the smoothness operator $L_p(F)$ can be seen as the limit of the eigenvectors for the graph Laplacian $L$ as the number of points goes to infinity. 

The eigenfunction calculation can be solved analytically for certain distributions. A numerical solution can be obtained by discretizing the density distribution into bins (centers and counts). Let $g$ be the eigenfunction values at a set of discrete points, then g satisfies:
\begin{equation}\label{eq:DiscrEfunc}
(\tilde{D}- P\tilde{W}P)g = \sigma P\hat{D}g 
\end{equation}
where $\sigma$ is the eigenvalue corresponding the eigenfunction $g$, $\tilde{W}$ is the affinity between the bins centers, P is a diagonal matrix whose diagonal elements give the density at the bins centers, $\tilde{D}$  is a diagonal matrix whose diagonal elements are the sum of the columns of $P\tilde{W}P$ , and $\hat{D}$ is a diagonal matrix whose diagonal elements are the sum of the columns of $P\tilde{W}$. The solution for eq. ~\ref{eq:DiscrEfunc} will be a generalized eigenvector problem of size $b\times b$, where $b$ is the number of discrete points of the density. Since $b<<n$, there is no need to construct the graph Laplacian matrix $L$ or solve for a more expensive generalized eigenvector problem. 

\begin{figure*}

\begin{tabularx}{\linewidth}{c|c|c|c|c}
    \hline
    \multirow{2}{*}[0.5in]{ \subfloat{\includegraphics[width=0.18\textwidth, height=0.15\textheight]{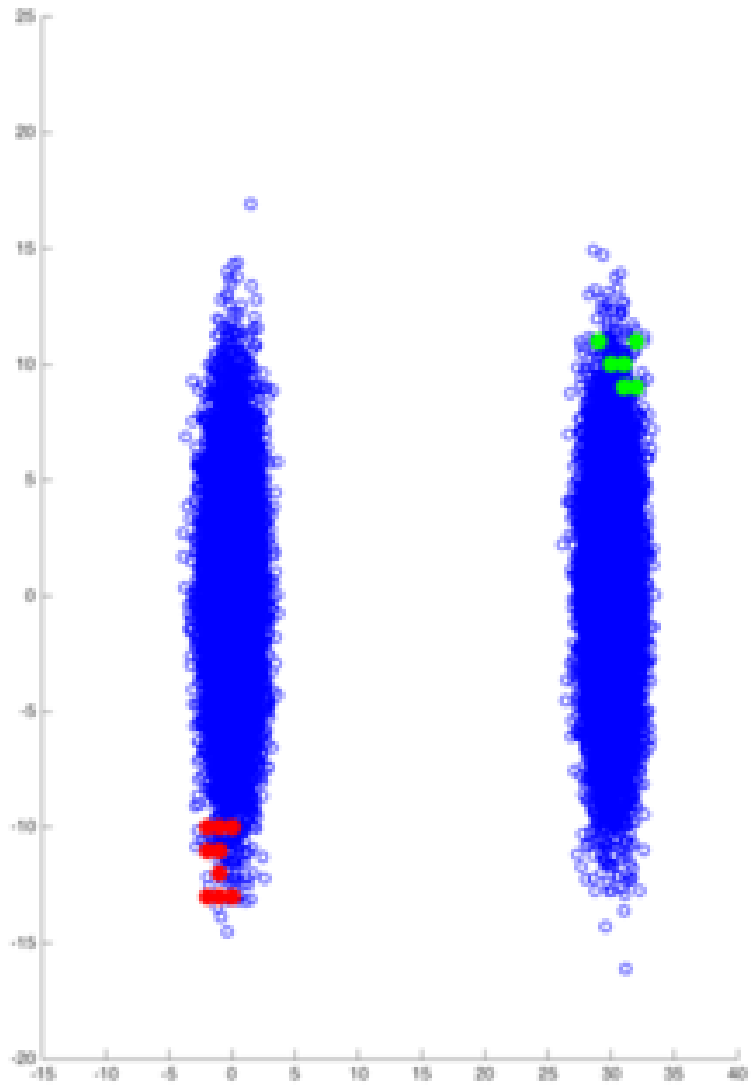}}}	&	\subfloat{\includegraphics[width=0.18\textwidth, height=0.10\textheight]{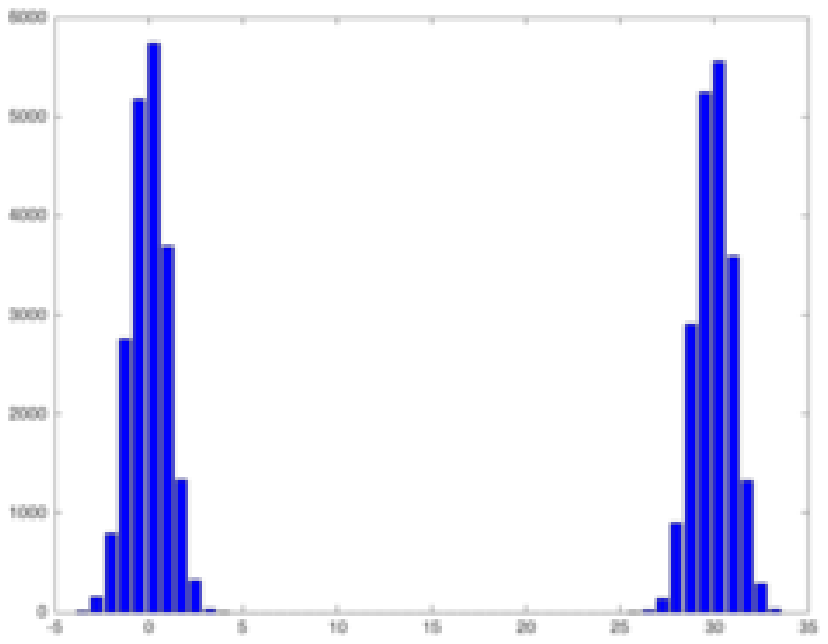}}	&	\subfloat{\includegraphics[width=0.18\textwidth, height=0.10\textheight]{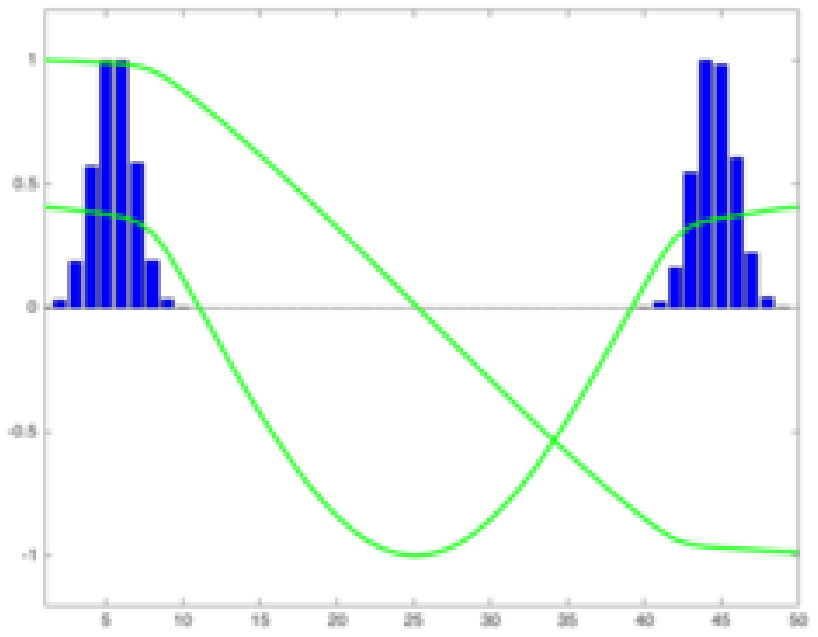}}	&		
\multirow{2}{*}[0.5in]{\subfloat{\includegraphics[width=0.18\textwidth, height=0.15\textheight]{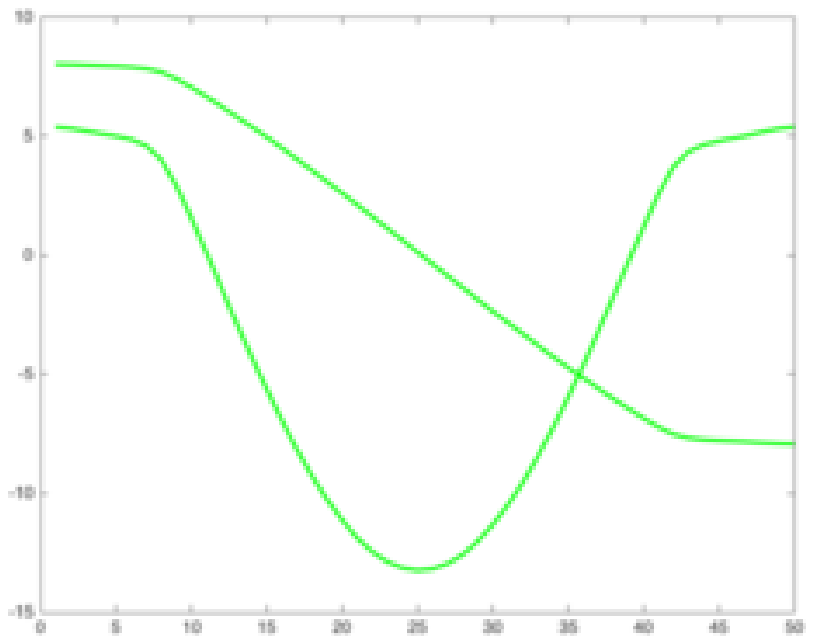}}} 
&	\multirow{2}{*}[0.5in]{\subfloat{\includegraphics[width=0.18\textwidth, height=0.15\textheight]{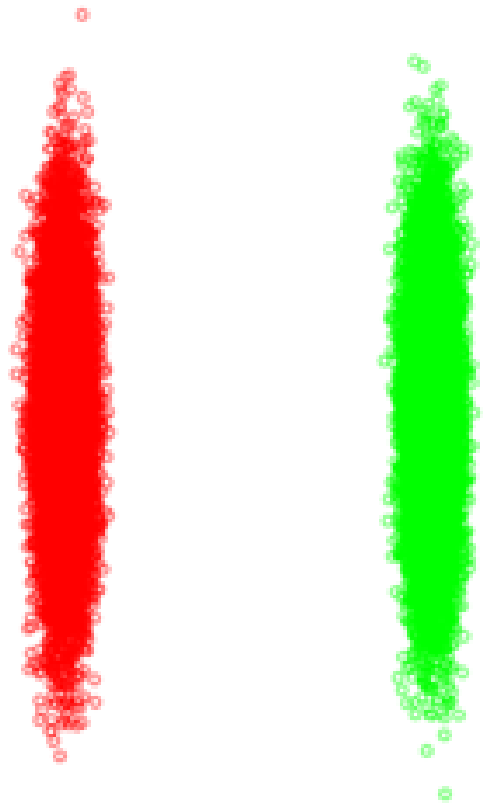}}}\\

    &\subfloat{\includegraphics[width=0.18\textwidth, height=0.10\textheight]{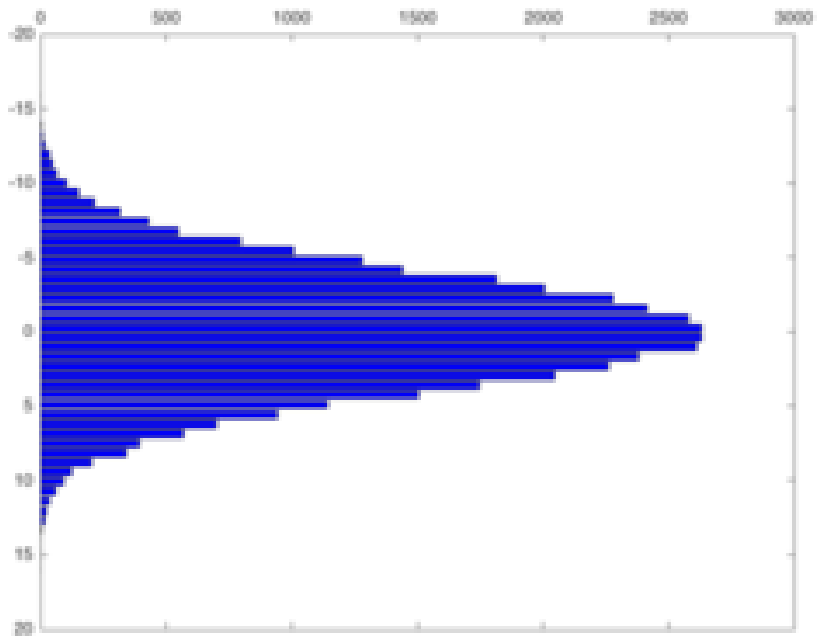}} & \subfloat{\includegraphics[width=0.18\textwidth, height=0.10\textheight]{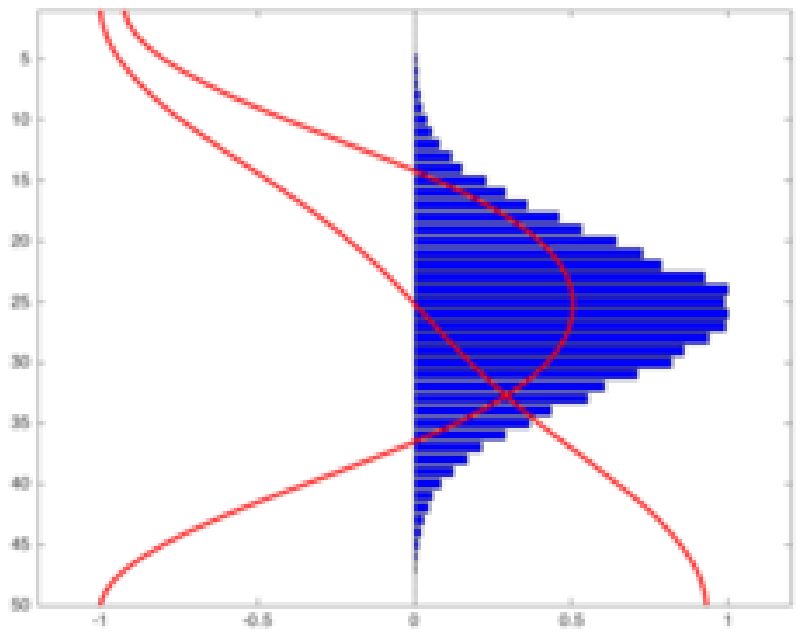}} &\\
    \hline
\end{tabularx}

        \caption{Semi-supervised learning using eigenfunctions. Column 1 shows a toy dataset for a semi-supervised learning problem. Column 2 shows the projection of dataset over its dimensions (X,Y) respectively. Then, eigenfunctions $g_i$ are calculated using Eq ~\ref{eq:DiscrEfunc}. Column 3 shows the first two eigenfunctions, with smallest non-zero eigenvalue, calculated for the data distribution over each dimension. Column 4 shows the two eigenfunctions with the smallest non-zero eigenvalues after aggregating eigenfunctions across the X and Y dimensions. Column 5 shows the classification result of the dataset based on the eigenvectors interpolated from the eigenfunctions.}
	\label{fig:why_it_works}
\end{figure*}

Figure ~\ref{fig:why_it_works} shows the eigenfunctions over separate dimension's distribution. The eigenfunctions with smallest eigenvalues are selected to solve the semi-supervised learning problem. Finally, the selected eigenfunctions are interpolated to calculate the eigenvectors of the Laplacian Matrix. For every eigenfunction calculated, a 1D interpolation is applied at the labelled points $x_l$.



%% file: Approach_review.tex
\section{Approach}\label{Sec:Approach}

To breakup our approach complexity, we divide it into three independent stages: feature extraction, Laplacian smoothness computation, and post processing. These stages are detailed in the following three subsections. 

\subsection{Feature Extraction}
\begin{figure}[h!]
	
	\centering\subfloat[Image]{\includegraphics[width=0.15\textwidth, height=0.10\textheight]{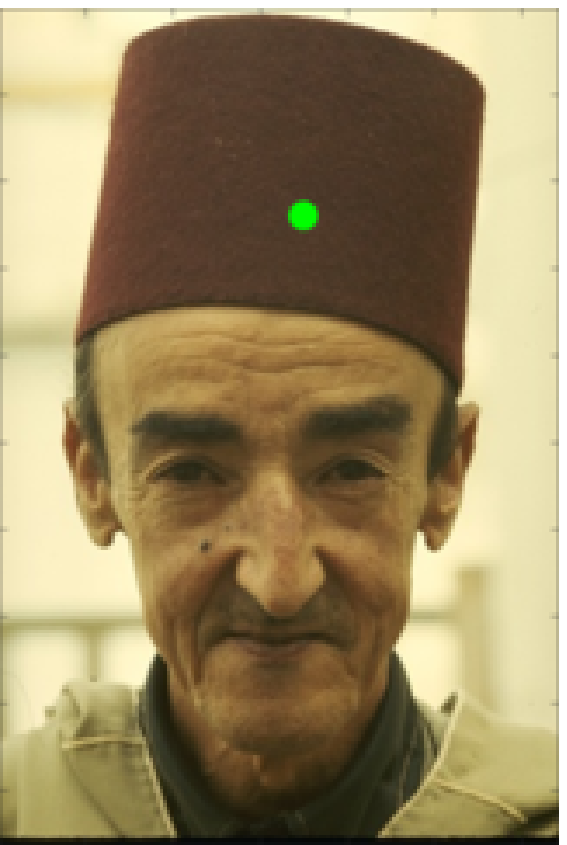}}    \hfil
	\centering\subfloat[RGB]{\includegraphics[width=0.15\textwidth, height=0.10\textheight]{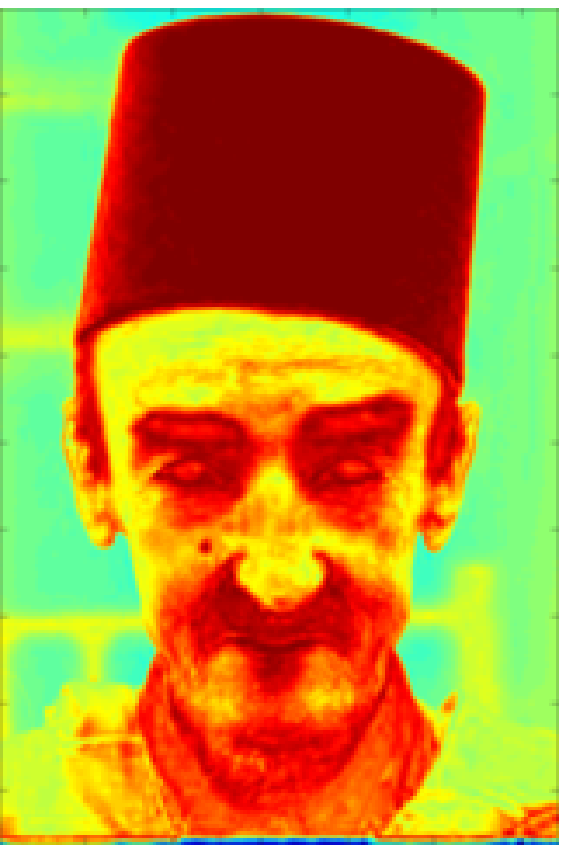}}    \hfil
	\centering\subfloat[LAB]{\includegraphics[width=0.15\textwidth, height=0.10\textheight]{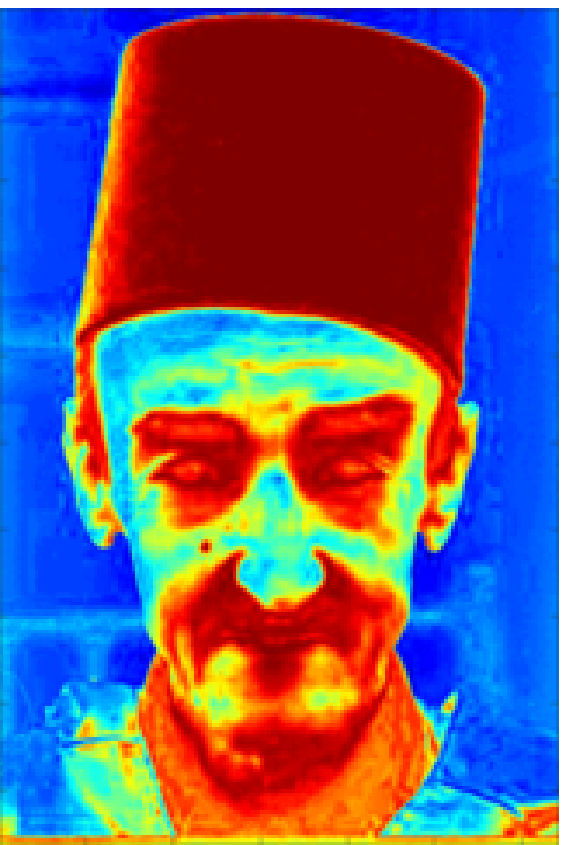}}    \hfil
	\subfloat[Euclidean]{\includegraphics[width=0.15\textwidth, height=0.10\textheight]{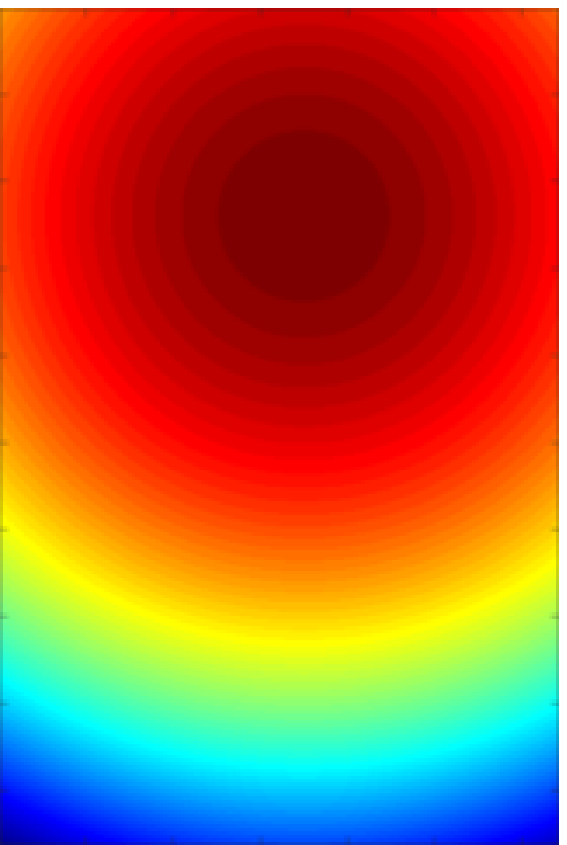}}    \hfil
	\centering\subfloat[Geodesic]{\includegraphics[width=0.15\textwidth, height=0.10\textheight]{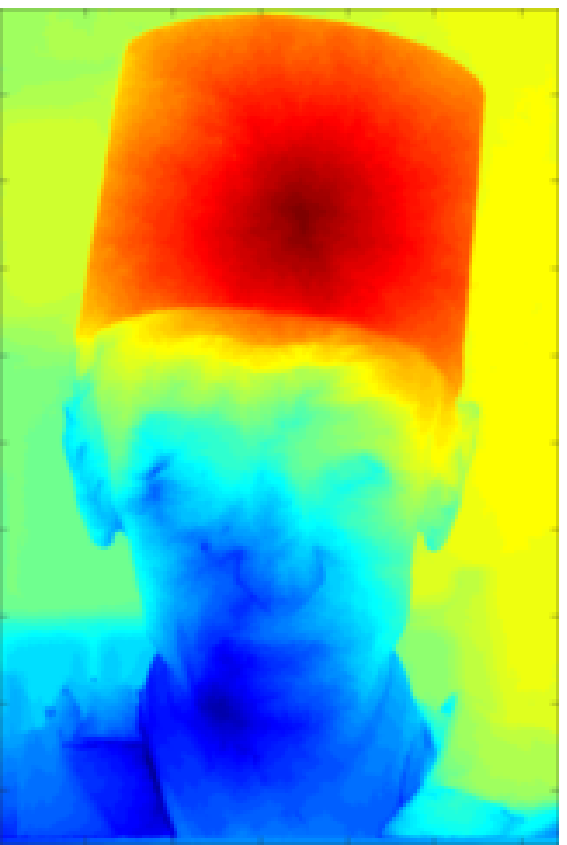}}    \hfil
	\centering\subfloat[IC]{\includegraphics[width=0.15\textwidth, height=0.10\textheight]{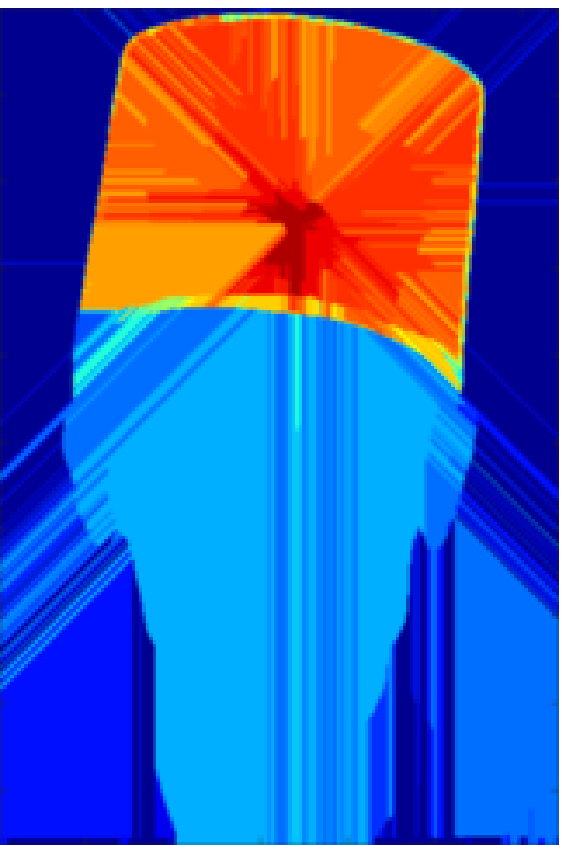}}    
	\caption{The first image shows the original RGB image with a green pivot. The rest of images show the pixel-to-pivot affinity results computed over RGB, LAB color model, euclidean, geodesic distance and intervening contour (IC) respectively. Best seen in color and zoom}\label{fig:Pixels_Features}
\end{figure}

Right features extraction is the most important stage in SL. Experiments with color spaces and geodesic features provide promising qualitative results. However, our control experiments show that these features do not generalize well on challenging datasets. So, alternative features combination are required. We evaluated different combinations of the five features shown in figure ~\ref{fig:Pixels_Features}: RGB, LAB, intervening contour, geodesic, and euclidean distance.

SL uses four out of the five features, so every pixel feature vector has length $4 \times (B + F)$, where $F$ and $B$ are the number of labeled foreground and background pixels. This vector encodes the affinity between an unlabeled and every labeled pixel using the four features. Such dependency on the number of labeled pixels ($F$, $B$) leads to a computational challenge. 
Thus, we propose to control the vector length by sampling pivots from the user scribble.

\begin{figure}[h!]
	\centering\subfloat[Human Annotation]{\includegraphics[width=0.3\textwidth]{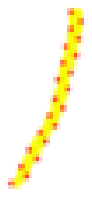}}    ~
	\centering\subfloat[Robot User Annotation]{\includegraphics[width=0.3\textwidth]{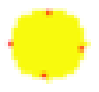}}    
	
	\caption[Sampling Approach]{We sample foreground and background scribbles to obtain a representative set of pivots. We sample the pivots uniformly from the enclosing contour.}\label{fig:sampling}
\end{figure}

We sample $k_1$ and $k_2$ pivots from foreground and background scribbles respectively. Thus, a pixel feature vector length is reduced to $4 \times (k_1 + k_2)$ as affinities are computed between pixels and pivots only. Pivots are sampled uniformly from the enclosing contour as shown in figure ~\ref{fig:sampling}. Such method is simple and does not degrade the accuracy of SL.

After pivots sampling, we compute the pixels feature vectors by measuring the pixel-to-pivot affinities. Initially, we build pixel-to-pivot affinities using the five pixel features: (1) RGB; (2) LAB color space; (3) Spatial proximity; (4) Intervening contour; and (5) Geodesic distance. 	After further investigation through control experiments, we decided to drop the use of intervening contour. Geodesic distance is a good substitute providing boundary cues for our segmentation procedure.

\begin{figure}[h!]
	\centering
	\subfloat[Feature Concatenation]{\includegraphics[width=0.3\textwidth,height=0.15\textheight]{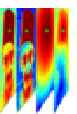}} ~
	\centering\subfloat[Feature Multiplication]{\includegraphics[width=0.3\textwidth,height=0.15\textheight]{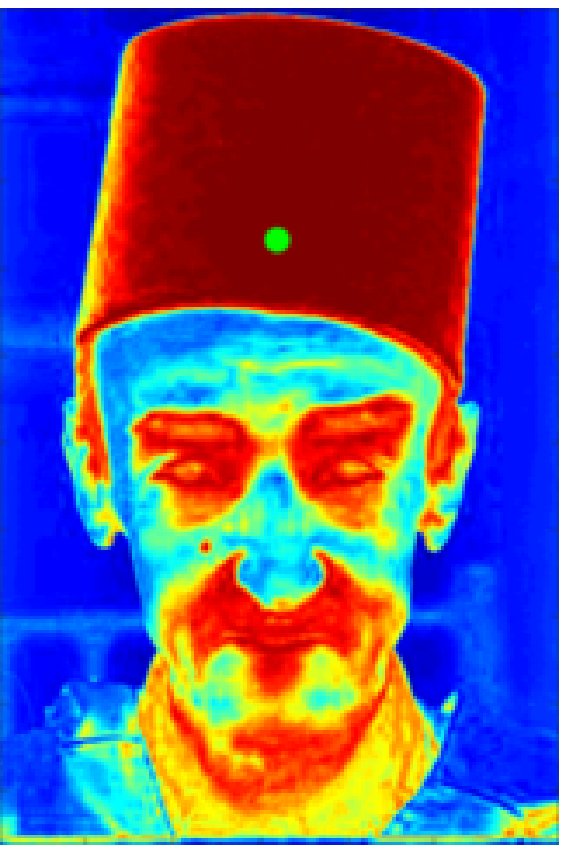}}                                 \caption{Different feature augmentation methods. These features are computed for one sample pivot centered inside the man's hat. Feature augmentation through multiplication is shown to be better in our control experiments}\label{fig:feature_augmentation}
\end{figure}

To decrease the vector dimensionality even further, we augment features using multiplication. Instead of stacking the features after each other, we multiply different affinities together. Thus, we end up with two alternatives to augment different pixel-to-pivot affinities' feature vectors as shown in figure ~\ref{fig:feature_augmentation}:

\textbf{Feature concatenation.} For every pixel, we compute a color affinity to the pivots, and do the same for the spatial proximity,  and geodesic affinities. This will end up with a vector of size $4 \times (k_1+k_2)$ for every pixel.

\textbf{Feature multiplication.} According to ~\cite{varma2009more}, the product kernels tend to produce better results for kernel combination in recognition problems. So, instead of concatenating the color, spatial, and geodesic affinities, we multiply them together. The resultant affinity vectors $k_1+k_2$ are concatenated with the original RGB and LAB color features for the pixel. This will result in a compact vector of size $k_1+k_2+6$ for every pixel. \\

Both feature augmentation alternatives are evaluated. Control experiments show that feature multiplication approach achieves better results.

After feature augmentation, we apply Principle Component Analysis (PCA) ~\cite{jolliffe1986introduction,smith2002tutorial} on the feature vectors to ensure data separability and dimensionality independence. 

Our experiments show that euclidean affinity is vital for higher accuracy. Unfortunately, it suppresses other features affinities. Such observation explains why this feature is not commonly used in segmentation approaches despite being intuitive. We apply regular normalization technique to limit such unfavorable effect. The euclidean feature is divide by its variance. To cope with different objects sizes, we multiply the variance by multiple scales, compute the Laplacian smoothness for each scale independently. Finally, we average all the Laplacian smoothness together.  
 

Algorithm ~\ref{algo:stage_1} summarizes the feature extraction stage details. 
\begin{algorithm}[H]
	\scriptsize
	\begin{algorithmic}
		\caption{Feature Extraction Stage}
		\label{algo:stage_1}
		\State Sample $k_1$ foreground and $k_2$ background pivots from user annotations [fig \ref{fig:sampling}]
		\State Compute pixel-to-pivots affinity feature vectors for every pixel.
		\State Normalize the Euclidean affinity feature by its variance.
		\State Augment pixel-to-pivot affinity feature vectors by multiplication.
		\State Apply PCA to the feature vectors.
\end{algorithmic}
\end{algorithm}

\subsection{Laplacian smoothness}
After applying PCA on feature vectors, we compute the Laplcian smoothness. First, we build a histogram to approximate the probability density function (PDF) of each independent dimension.

Given the approximate density for each dimension, we solve numerically for eigenfunctions $g$ and eigenvalues $\sigma$ using eq. ~\ref{eq:DiscrEfunc}. The eigenfunctions from all dimensions are sorted by increasing eigenvalue. The $m$ eigenfunctions with the smallest eigenvalues are selected. Now we have $m$ functions ${ \Phi  }_{ k }$ whose values are given at a set of discrete points for each coordinate. 1D Linear interpolation is used to interpolate ${ \Phi  }_{ k }$ at each of the labeled points $x_l$ to compute Laplacian eigenvectors $U$. The Laplacian smoothness is measured by $f = U\alpha$ (eq. \ref{eq:laplacian_smoothness_equation}) where $U$ is a $n \times m$ matrix whose columns are the $m$ eigenvectors with smallest eigenvalues. This allows us to solve eq. ~\ref{eq:solve} in a time complexity that is independent of the  unlabeled points number. 

Despite being similar to Fergus et. al approach ~\cite{fergus2009semi}, direct implementation would suffer high latency due to operations on large matrices. To better serve the nature of our problem, slow operations are optimized. We substitute matrices operations by equivalent vector ones that scale much better as the number of unlabeled points increases. Algorithm ~\ref{algo:stage_2} summarizes the steps for computing Laplacian smoothness.

\begin{algorithm}[H]
	\scriptsize
	\begin{algorithmic}
		\caption{Compute Laplacian Smoothness Stage}
		\label{algo:stage_2}
		\State Compute Laplacian eigen-functions from feature vectors
		\State \hspace{\algorithmicindent} Build a histogram to approximate the PDF of each independent dimension.
		\State \hspace{\algorithmicindent} Solve numerically for eigenfunctions and eigenvalues using eq. ~\ref{eq:DiscrEfunc}. 
		\State \hspace{\algorithmicindent} Sort eigenfunctions ascendingly
		\State \hspace{\algorithmicindent} Select $m$ eigenfunctions with smallest eigenvalues $\alpha$.
		\State Interpolate eigen-vectors $U$ from  eigen-functions
		\State For each euclidean affinity normalized at different scales s, compute Laplacian smoothness $ f_s = U\alpha$
		\State Final Laplacian smoothness = $\sum_{s}{average(f_s)}$
	\end{algorithmic}
\end{algorithm}

\subsection{Post processing}
The final post processing stage is straight forward. The final Laplacian smoothness is zero thresholded. Assign foreground and background labels to +ve and -ve values. Finally, remove small islands and fill small holes in the segmented object.

\subsection{Single Pass}

The three-stages approach presented is used to process a query image when all annotations are available beforehand. Thus, it is called \textit{single pass}. In a real scenario, users want to refine the segmentation result by interactively providing more annotations. To adapt such behavior, a robot user variant is presented that can incrementally handle new annotations.

\subsection{Robot User}
To mimic human behavior, we use a robot user. The robot user generates a flexible sequence of user interactions, according to well-defined rules, that model the way in which residual error in segmentation is progressively reduced in an interactive system. We use the standard deviation of the error progressively to assess the reliability of the model. 

An ideal evaluation system would measure the amount of effort required to reach a certain band of segmentation accuracy. Thus, the robot user simulates user interactions by placing brushes automatically. Initially, Seeded Laplacian starts with an initial set of brush strokes (chosen manually with one stroke for foreground and three strokes for background) and computes a segmentation. Then the robot user places a circular brush stroke with diameter 17 pixels in the largest connected component of the segmentation error area, placed at a point farthest from the boundary of the component. The process is repeated up to 20 times, generating a sequence of 20 simulated user strokes. Further demonstration and details for robot user is available in ~\cite{nickisch2010learning}.

Two Seeded Laplacian variants were developed to handle robot user annotations: the naive approach and the incremental approach.

\textbf{Naive approach}: We calculate the whole new solution every time the robot user adds an annotation. 

\textbf{Incremental approach}: We sample the pivots from the last annotation only. Then, Seeded Laplacian computes the pixel-to-pivots affinity features with respect to this new annotation. These new feature vectors are concatenated with the previously computed one. Then, we apply the same eigenfunction and eigenvector calculation procedure as in the single pass approach over the newly augmented feature vectors. 

%% file: experimental_results_review.tex
\section{EXPERIMENTAL RESULTS}
\subsection{Time Complexity Analysis}
\begin{figure}[h!]
	\centering
        \subfloat[ Toy Dataset]{\includegraphics[width=0.4\textwidth]{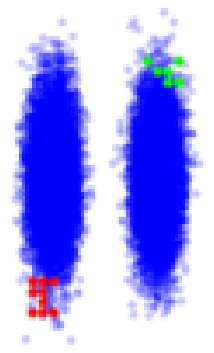}}    
          \centering
          \subfloat[Semi-Supervised classification]{\includegraphics[width=0.4\textwidth]{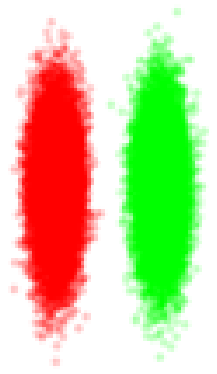}}    
          
\begin{tikzpicture}
\begin{axis}[ 
xlabel={No. of Points},
  ylabel={Calculation Time (secs)},
legend entries={Eig. vector,Eig. fuction ~\cite{fergus2009semi},\textbf{Our} Eig function opt}] 
\addplot coordinates {  (400,0.41)  (900,2.33) (1600,12) (2500,45) (3600,151) }; 
\addplot coordinates {  (400,0.2) (900,0.29) (1600,0.57) (2500,0.75) (3600,0.96)(4900,0.68) (6400,1.03)   (15000,5.7) (20000,26) (30000,67) }; 
\addplot coordinates {  (400,0.04) (900,0.08) (1600,0.04) (2500,0.11) (3600,0.15)(4900,0.03) (6400,0.04)  (15000,0.03) (20000,0.07) (30000,0.09) }; 
\end{axis} \end{tikzpicture}
\caption{Time Complexity Analysis. Comparing different approaches for computing laplacian eigenvectors. Our optimized version copes well as the number of points increases}
\label{fig:time_analysis}
\end{figure}

Due to our problem interactive nature, Laplacian smoothness procedure had to speed up. So, first we present a brief time complexity analysis in figure ~\ref{fig:time_analysis}. We sample points from two Gaussian distributions and solve for the semi-supervised classification. We vary the number of samples and measure the time needed to compute the eigenfunction approximate solution of eq.~\ref{eq:DiscrEfunc} versus the eigenvector solution of eq.~\ref{eq:solve}. We developed an optimized version of the eigenfunction solution that we call eigenfunction optimized. The main difference between our implementation versus the implementation of ~\cite{fergus2009semi} is that we vectorize most of the matrix operations in their code. 

Deviating from the implementation of ~\cite{fergus2009semi}, our optimized version \footnote{Github Source: \url{https://github.com/ahmdtaha/ssl_opt}} reduces the computational cost of the $\Lambda { U }$ multiplication operation by some simple scalar multiplications. We define ${U}_{labeled}$ as a sub-matrix of ${ U }$ containing the rows corresponding to the labeled pixels. Then ${U}_{labeled}$ is multiplied by $\lambda$ scalar value as $\lambda {U}_{labeled}$. A new zero matrix of  $size(\Lambda { U })$ is constructed, and the result of $\lambda {U}_{labeled}$  is inserted into the zero matrix. A similar approach is applied to reduce the $\mathbf{\Lambda y}$ computational cost.
\subsection{Comparative Evaluation}
\textbf{Datasets:} One of the main challenges for developing a scribble-based segmentation approach is the evaluation process. Every new scribble segmentation approach develops its own evaluation dataset. That makes it difficult to evaluate different approaches without being biased toward a particular method or dataset.

One contribution in this paper is the construction of a 700 annotated image segmentation dataset. Our goal is to provide a standard evaluation procedure and quantitative benchmark for different scribble-based segmentation approaches. To annotate such large image collection without being biased to a particular segmentation approach, we outsourced this task.

\begin{figure*}
\begin{tabularx}{\linewidth}{cccc}
	\multirow{2}{*}[0.5in]{ \subfloat{\includegraphics[width=0.3\textwidth, height=0.15\textheight]{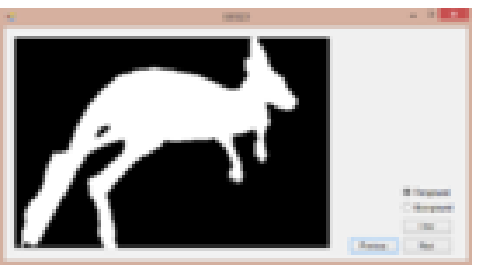}}}	&	\subfloat{\includegraphics[width=0.18\textwidth, height=0.10\textheight]{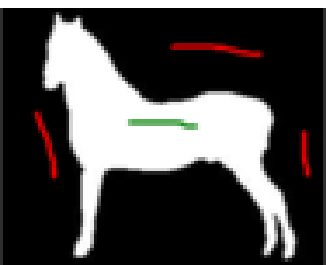}}	&	\subfloat{\includegraphics[width=0.18\textwidth, height=0.10\textheight]{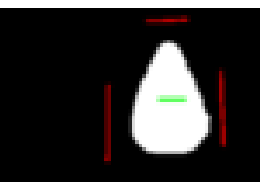}}	&		
	\subfloat{\includegraphics[width=0.18\textwidth, height=0.10\textheight]{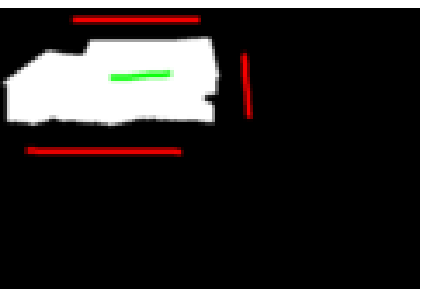}} \\

	&\subfloat{\includegraphics[width=0.18\textwidth, height=0.10\textheight]{horse004}} & \subfloat{\includegraphics[width=0.18\textwidth, height=0.10\textheight]{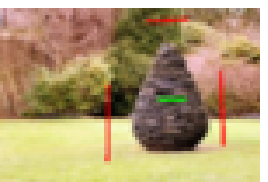}} &
	\subfloat{\includegraphics[width=0.18\textwidth, height=0.10\textheight]{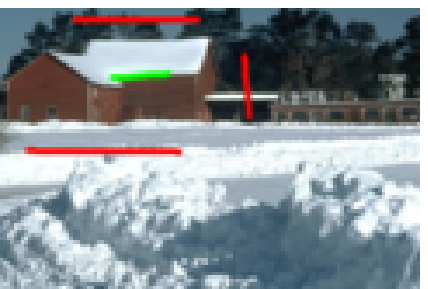}} \\
\end{tabularx}
\caption{Creating annotation for new datasets. Left image shows the software used by annotators to create scribble annotations. Annotators use mouse inside action to draw the scribbles. Top row shows the annotators' view. Images are from Weizmann horses, Weizmann single object, and BSD 100 datasets. The bottom row shows the final annotations.}\label{fig:annotation_software}

\end{figure*}

Following ~\cite{gulshan2010geodesic} annotation style, every image has four scribbles divided as one foreground and three background scribbles. We created a software program to help generate the user scribbles for other well known interactive image segmentation datasets like Weizmann horses, BSD 100, Weizmann single, and two objects datasets. The software presents the annotator with the image's ground-truth as shown in figure ~\ref{fig:annotation_software}. 


\textbf{The Geodesic Star-Dataset}~\cite{gulshan2010geodesic} is a scribble-based interactive image segmentation dataset. The dataset consists of 151 images: 49 images taken from the GrabCut dataset, 99 from the PASCAL VOC dataset, and 3 images from the Alpha matting dataset. 

\textbf{The Weizmann horses dataset} ~\cite{borenstein2008combined} is a top-down and bottom-up segmentation dataset. The dataset contains 328 horse images that were collected from the various websites. The dataset foreground/background ground truth are manually segmented. The images are highly challenging; they include horses in different positions, such as running, standing, and eating. The horses also have different textures, e.g., zebra-like horses. The images’ background underlay varying amount of occlusion and lighting conditions. 

\textbf{The BSD 100 dataset} ~\cite{mcguinness2010comparative} consists of 100  distinct objects from publicly available datasets. 96 images were selected from the  Berkeley Segmentation 300 Dataset ~\cite{martin2001database}. Selected images represent a large variety of segmentation challenges, such as texture, cluttering, camouflage, and various lighting conditions. The ground truth is constructed by hand for better accuracy.

\textbf{Weizmann single and two objects dataset} ~\cite{alpert2007image}  consists of 200 images. The database is designed to contain a variety  of  images  with  objects  that  differ  from  their surroundings by either intensity, texture, or other low-level  cues. The dataset is divided into two portions: the single object set and the two objects set. In the single object set, 100 images are selected that clearly contain one foreground object. In the two objects set, another 100 images  are selected that contain two similar foreground  objects.

\noindent\textbf{Evaluation Measures:} In our experiments, we use 1) F-score and 2) Jaccard Index indexes as evaluation measures

\textbf{F-score} corresponds to the harmonic mean of positive predictive value (PPV) and true positive rate (TPR) ~\cite{witten2005data}; therefore, it is class-specific and symmetric. 
\begin{equation}
F_s = \frac { 2*TP }{ 2*TP+FP+FN } 
\end{equation}
where true positives (TP) and false positives (FP) are instances correctly and incorrectly classified, whereas true negatives (TN) and false negatives (FN) are instances correctly and incorrectly not classified. It can be interpreted as a measure of overlapping between the true and estimated classes.

\textbf{Jaccard index} (also known as overlap score) is initially defined to compare sets ~\cite{jaccard1912distribution}. It is a class-specific symmetric measure defined as:
\begin{equation}
JI = \frac { TP }{ TP+FP+FN } 
\end{equation}
For a given class, it can be interpreted as the ratio of the estimated and true classes intersection to their union in terms of set cardinality. It is linearly related to the F-measure such that $JI = F_s/(2-F_s)$.\\

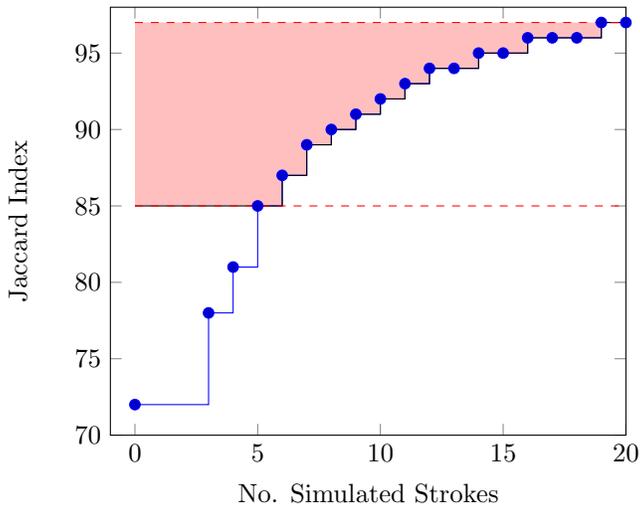
\begin{figure}[h!]
\centering
\begin{tikzpicture}
\begin{axis}[
xmin = -1,
xmax = 20,
ymin = 70,
ymax= 98,
xlabel={No. Simulated Strokes},
 ylabel={Jaccard Index}]

\addplot +[const plot,name path=C] coordinates {  
(0,72)
(3,78)
(4,81)
(5,85)
(6,87)
(7,89)
(8,90)
(9,91)
(10,92)
(11,93)
(12,94)
(13,94)
(14,95)
(15,95)
(16,96)
(17,96)
(18,96)
(19,97)
(20,97)
 }; 

\addplot [const plot,name path=D] coordinates {  
(0,85)  
(1,85)
(2,85)
(3,85)
(4,85)
(5,85)
(6,87)
(7,89)
(8,90)
(9,91)
(10,92)
(11,93)
(12,94)
(13,94)
(14,95)
(15,95)
(16,96)
(17,96)
(18,96)
(19,97)
(20,97)
 };

\addplot +[red,dashed,name path=B,mark=none] coordinates {  
(0,85)  
(20,85)
 };

\addplot +[red,dashed,name path=A,mark=none] coordinates {  
(0,97)  
(20,97)
 };

\addplot[pink] fill between[of=D and A];

\end{axis} \end{tikzpicture}
\caption[Robot User Effort Evaluation.]{Plotting overlap score against no. of strokes to measure interaction effort. The area above the curve is a measure of the average number of strokes required. Since we are interested in the amount of interaction required to achieve high segmentation accuracy, the average is restricted to the band [$Alow$, $Ahigh$], as illustrated (shaded in pink).
}\label{fig:robot_user_evaluation}
\end{figure}

\textbf{Average number of robot user strokes}: Interactive system quality is evaluated using the average number of strokes required to achieve segmentation quality within a certain band. This is illustrated in figure ~\ref{fig:robot_user_evaluation}. The graph of overlap score against the number of brush strokes captures how the segmentation accuracy increases with successive user interactions, and the average number of strokes summarizes that in a single score. The average is computed over a certain band of accuracy, and we take $A_{low} = 0.85$, $A_{high} = 0.98$. The average number of strokes is computed using the formulate as follow 
\begin{equation}
\textit{Avg. No. Strokes} = \frac { \textit{Area shaded above curve} }{ (0.98 - 0.85) } 
\end{equation}

\subsection{Control Experiments}

In this subsection, we demonstrate three control experiments that find the best parameter settings for our approach. We conduct all  control experiments over Geodesic Star-Dataset~\cite{gulshan2010geodesic} and the original user scribbles provided by the dataset. 

\textbf{Experiment 1}: In this experiment, our goal is to find:
\begin{enumerate}
\item The best pixel-to-pivot affinity feature vectors that produce the best segmentation results. 
\item The best feature augmentation method.
\end{enumerate}

\begin{figure}[h!]
	\centering
\begin{tikzpicture}
\begin{axis}[
xbar,
 enlarge y limits=0.07,
xlabel={Jaccard Index},
legend style={at={(0.5,-0.18)},
	anchor=north,legend columns=-1},
symbolic y coords={Geo,Geo + Euc,RGB+LAB,RGB+LAB +Euc,RGB+LAB +Geo,RGB+LAB +Euc+Geo,RGB+LAB +Euc+IC},
ytick=data
]
   \addplot coordinates
{(0.50,Geo)(0.59,Geo + Euc)(0.54,RGB+LAB)(0.63,RGB+LAB +Euc)(0.56,RGB+LAB +Geo)(0.6,RGB+LAB +Euc+Geo) (0.62,RGB+LAB +Euc+IC) };

            \addplot coordinates
{(0.58,Geo + Euc)(0.55,RGB+LAB)(0.68,RGB+LAB +Euc) (0.63,RGB+LAB +Geo)(0.69,RGB+LAB +Euc+Geo) (0.65,RGB+LAB +Euc+IC)};

\legend{Feature Concatenation,Feature Multiplication}

\end{axis}
\end{tikzpicture}
\caption[Control Experiments 1 findings]{Control Experiments 1. We investigate the best features for image segmentation purpose and the best augmentation method for these features. We conclude that feature augmentation through multiplication is better than concatenation. We also conclude that RGB+LAB+Euclidean affinity+Geodesic affinity are the best features to use.} \label{fig:control_experiments_1} 

\end{figure}
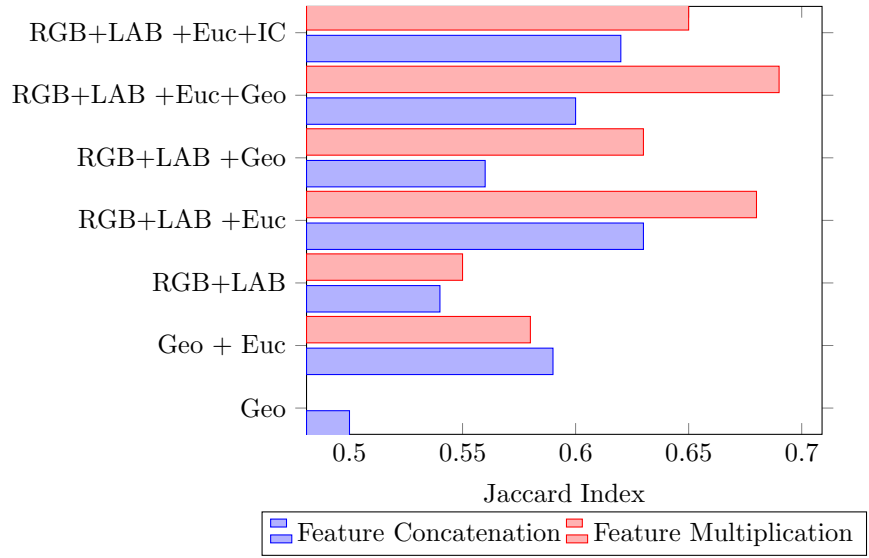

We compare feature augmentation alternatives; feature concatenation versus feature multiplication. From the results shown in figure ~\ref{fig:control_experiments_1}, it is clear that feature augmentation through multiplication provides superior results over feature concatenation. The best affinity features to use for image segmentation is another insight we gain from the same experiments.\\
The improvement caused by adding the spatial feature vector (Euclidean affinity) is significant. We find RGB+LAB feature vectors to be indispensable. To include boundary cues in our approach, we examined both geodesic affinity and intervening contour. Geodesic affinity reported better results. The final set of affinity features employed in our approach are RGB+LAB+Spatial (Euc) + Geodesic.

\textbf{Experiment 2}: Our aim in this experiment is to find the best number of eigenvectors that achieve best segmentation results. We used the settings concluded from control experiment no.1. Affinity feature vectors are RGB+LAB+Spatial (Euc) + Geodesic. These features are augmented by multiplication.
 
\begin{figure}[h!]
\centering
\begin{tikzpicture} 
\begin{axis}[
xlabel={Number of eigenvectors},
ylabel={Jaccard Index},
width=0.5\textwidth,
legend pos=south east,
 symbolic x coords={25,50,75,100,125,150},
xtick=data,
ybar,
]
    \addplot   coordinates
        {(25,0.61)(50,0.684)(75,0.686)(100,0.682)(125,0.682) (150,0.682) };
\end{axis}
\end{tikzpicture}
\begin{tikzpicture} 
\begin{axis}[
xlabel={Number of pivots},
ymin=0.66,
ymax=0.69,
width=0.5\textwidth,
legend pos=south east,
symbolic x coords={6,9,21,42},
xtick=data,
ybar,
]
\addplot coordinates
{(6,0.6733)(9,0.6821)(21,0.6824)(42,0.6841) };
\end{axis}

\end{tikzpicture}
\caption[Control Experiments 2 findings]{Control Experiments 2 (Left) and 3 (Right). We investigate the best number of eigenvectors and pivots for computing laplacian smoothness. From experiment no.2,  we conclude that 100 eigenvectors are best for our approach. Such number will cope well when the image size increases. It can also work with large number of foreground and background pivots. Experiment no.3 shows that 42 pivots from foreground and background scribbles, 21 from each, provide best performance.} \label{fig:control_experiments_2} 
\end{figure}
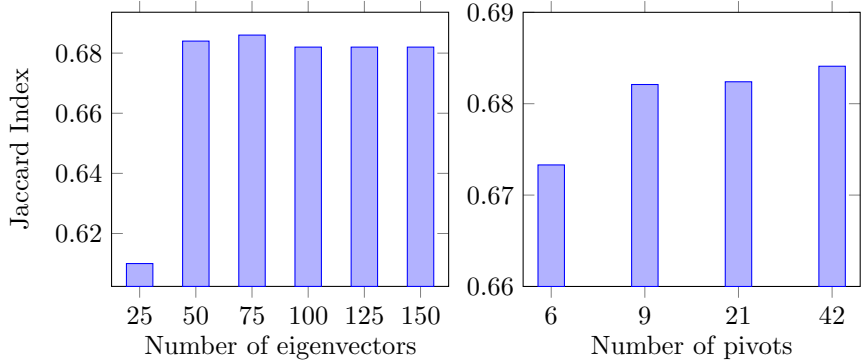

From figure ~\ref{fig:control_experiments_2} (Left), we conclude that the number of eigenvectors used does not improve the accuracy significantly after 50 eigenvectors. We decided to use 100 eigenvectors for a number of reasons. Firstly, we want to compromise between the speed of our approach and its accuracy.  Secondly, the standard deviation of the 100 eigenvectors segmentation result is less than their corresponding 50 and 75 eigenvectors. Finally, 100 eigenvectors can cope with possible increase in the number of feature vectors. The number of feature vectors increase if the number of foreground/background pivots increases.

\textbf{Experiment 3}: In this experiment, our goal is to find the best number of pivots to sample from user scribbles. We use the same parameter settings concluded from the previous control experiments and study the effect of changing the number of pivots on the segmentation accuracy. From figure ~\ref{fig:control_experiments_2} (Right), we decided to use 21 pivots.

\subsection{Quantitative Evaluation}

We quantitatively compare our proposed method with multiple algorithms, including BJ, RW, PP, SP-IG, SP-LIG, and SP-SIG ~\cite{boykov2001interactive} ~\cite{grady2006random} ~\cite{liu2009paint}~\cite{gulshan2010geodesic} which gives best performance on scribble segmentation reported by ~\cite{gulshan2010geodesic}. GSCseq and ESC are demonstrated as state-of-the-art ~\cite{gulshan2010geodesic}. In all experiments, we set the number of eigenfunctions to 100, the number of foreground pivots to 21, and the number of background pivots to 21.

The performance of various scribble segmentation algorithms are evaluated using Geodesic Star-Dataset annotations. In this experiment, each algorithm is presented by an annotation image that contain one scribble as foreground and three other scribbles as background. Table ~\ref{table:1} shows a detailed comparison between SL and other segmentation approaches. It is clear that SL outperforms all other segmentation methods. We used standard deviation measure to study the stability of the segmentation approaches. SL is very competitive with state-of-the-art methods, and SL's standard deviation is superior to most segmentation methods.

\begin{table}
\centering
\caption{Geodesic Star-Dataset Comparative Evaluation Experiment.}
\begin{tabular}{|l|c|c|}
\hline
Method Name & Jaccard Index & F Score  \\ \hline
BJ          & 0.49 $\pm$ 0.26            &  0.62 $\pm$ 0.23      \\ \hline
RW        & 0.53 $\pm$ 0.21            &  0.67 $\pm$ 0.18      \\ \hline
PP          & 0.59  $\pm$  0.25           &     0.70 $\pm$ 0.21  \\ \hline
GSC      & 0.61  $\pm$ 0.25           &      0.72 $\pm$ 0.21    \\ \hline
ESC         & 0.61 $\pm$ 0.24            &    0.72 $\pm$ 0.21  \\ \hline
SP-IG       & 0.56 $\pm$  0.16            &   0.70 $\pm$ 0.13       \\ \hline
SP-LIG         & 0.59 $\pm$ 0.22            &    0.72 $\pm$ 0.18  \\ \hline
SP-SIG         & 0.62 $\pm$ 0.17            &    0.75 $\pm$ 0.13     \\ \hline \hline
\textbf{SL}            & \textbf{0.69} $\pm$ 0.17            &    \textbf{0.80}  $\pm$ 0.13   \\ \hline
\end{tabular}
\label{table:1}
\end{table}

\subsection{Qualitative Evaluation}
\begin{figure*}

	\subfloat{\includegraphics[width=0.15\textwidth, height=0.08\textheight]{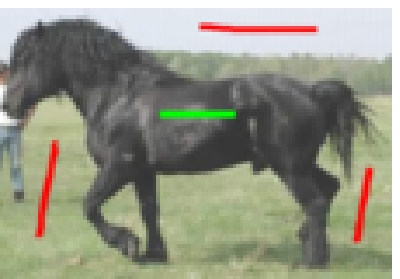}}   \hfill
	\subfloat{\includegraphics[width=0.15\textwidth, height=0.08\textheight]{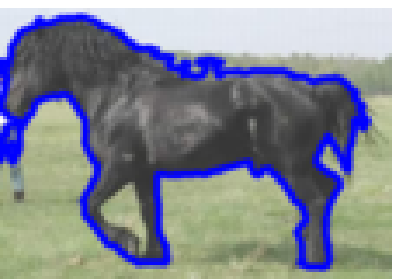}}    \hfill
	\subfloat{\includegraphics[width=0.15\textwidth, height=0.08\textheight]{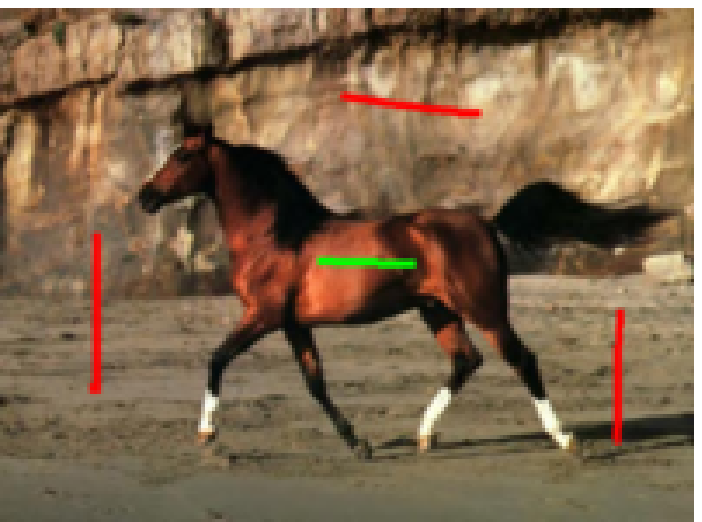}}    \hfill
	\subfloat{\includegraphics[width=0.15\textwidth, height=0.08\textheight]{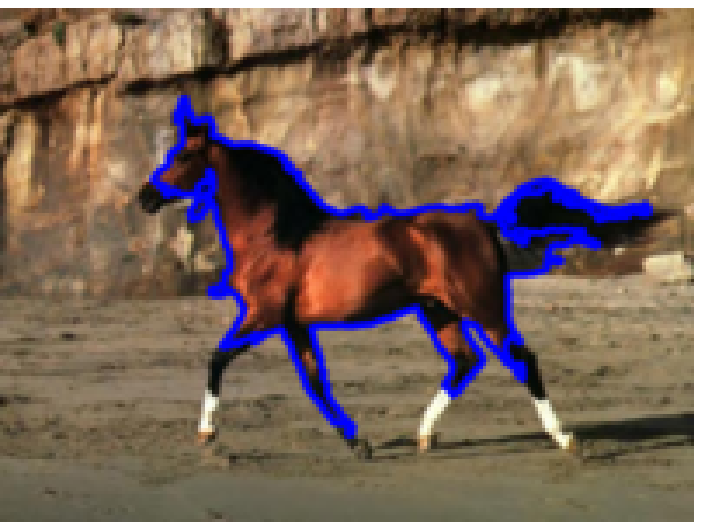}}  \hfill
	\subfloat{\includegraphics[width=0.15\textwidth, height=0.08\textheight]{horse004}}    \hfill
	\subfloat{\includegraphics[width=0.15\textwidth, height=0.08\textheight]{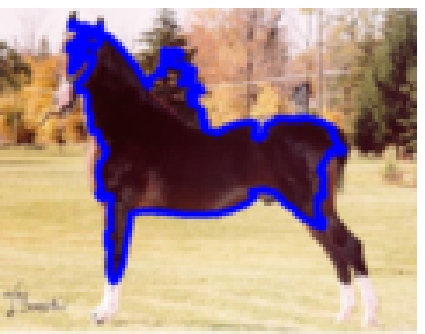}} \hfill
	\vspace{1ex}	

	\subfloat{\includegraphics[width=0.15\textwidth, height=0.08\textheight]{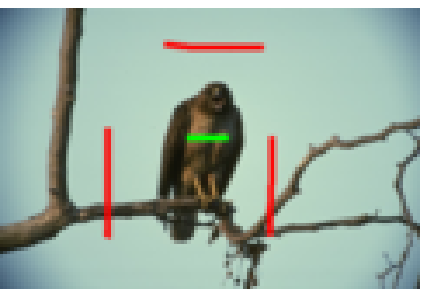}}   \hfill
	\subfloat{\includegraphics[width=0.15\textwidth, height=0.08\textheight]{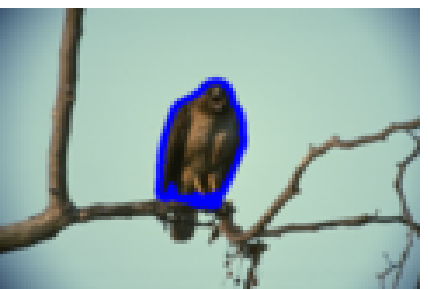}}    \hfill
	\subfloat{\includegraphics[width=0.15\textwidth, height=0.08\textheight]{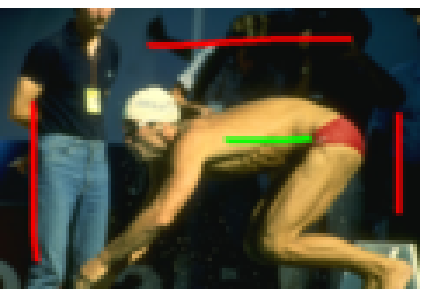}}    \hfill
	\subfloat{\includegraphics[width=0.15\textwidth, height=0.08\textheight]{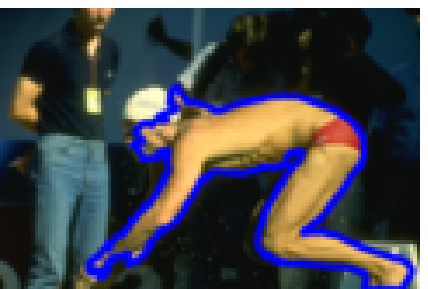}}  \hfill
	\subfloat{\includegraphics[width=0.15\textwidth, height=0.08\textheight]{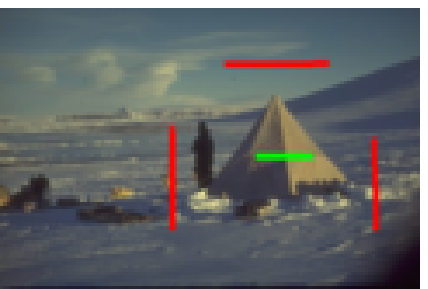}}    \hfill
	\subfloat{\includegraphics[width=0.15\textwidth, height=0.08\textheight]{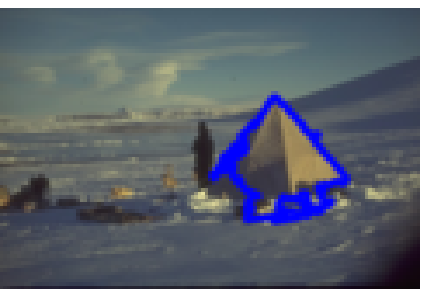}} \hfill
	\vspace{1ex}	
	\subfloat{\includegraphics[width=0.15\textwidth, height=0.08\textheight]{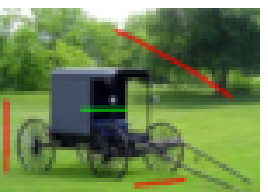}}   \hfill
	\subfloat{\includegraphics[width=0.15\textwidth, height=0.08\textheight]{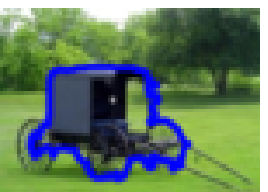}}    \hfill
	\subfloat{\includegraphics[width=0.15\textwidth, height=0.08\textheight]{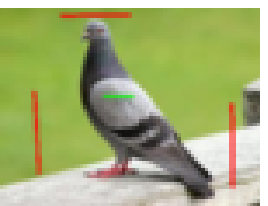}}    \hfill
	\subfloat{\includegraphics[width=0.15\textwidth, height=0.08\textheight]{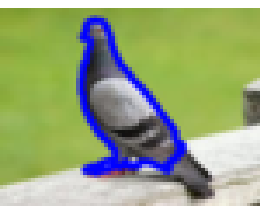}}  \hfill
	\subfloat{\includegraphics[width=0.15\textwidth, height=0.08\textheight]{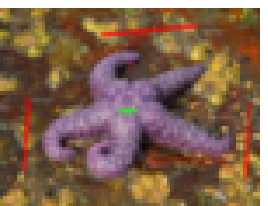}}    \hfill
	\subfloat{\includegraphics[width=0.15\textwidth, height=0.08\textheight]{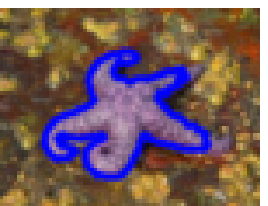}} \hfill
	\vspace{1ex}	
	\subfloat{\includegraphics[width=0.15\textwidth, height=0.08\textheight]{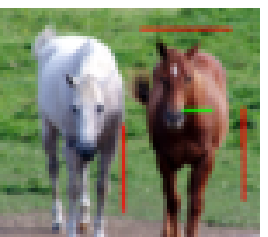}}   \hfill
	\subfloat{\includegraphics[width=0.15\textwidth, height=0.08\textheight]{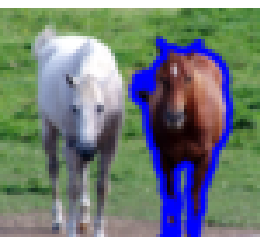}}    \hfill
	\subfloat{\includegraphics[width=0.15\textwidth, height=0.08\textheight]{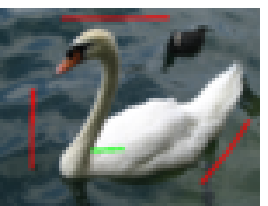}}    \hfill
	\subfloat{\includegraphics[width=0.15\textwidth, height=0.08\textheight]{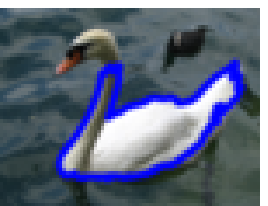}}  \hfill
	\subfloat{\includegraphics[width=0.15\textwidth, height=0.08\textheight]{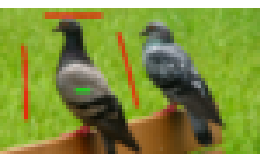}}    \hfill
	\subfloat{\includegraphics[width=0.15\textwidth, height=0.08\textheight]{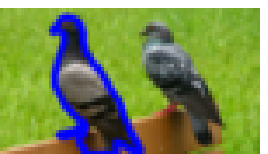}} \hfill
	\vspace{1ex}			

        \caption{SL Qualitative results across four different scribble-based interactive image segmentation datasets. Every row contains images from different dataset. The datasets appear in the following order: Weizmann Horses, BSD 100, Weizmann single object, and Weizmann two objects datasets.   }\label{fig:qualitative_results_geodesic}
\end{figure*}

Figure ~\ref{fig:qualitative_results_geodesic} shows the qualitative results of SL over four different datasets. The combination of spatial proximity, geodesic distance, and different color models enables SL to grab large region of the foreground object while being sensitive to edges. 

\begin{figure}

	\subfloat{\includegraphics[width=0.15\textwidth, height=0.10\textwidth]{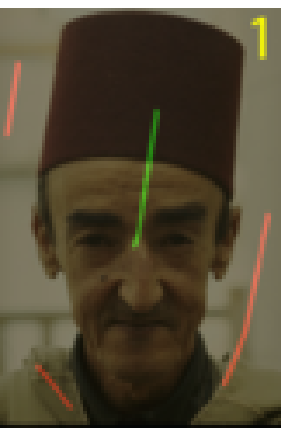}}   \hfill
		\subfloat{\includegraphics[width=0.15\textwidth, height=0.10\textwidth]{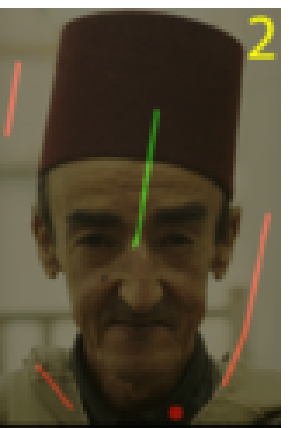}}   \hfill
	\subfloat{\includegraphics[width=0.15\textwidth, height=0.10\textwidth]{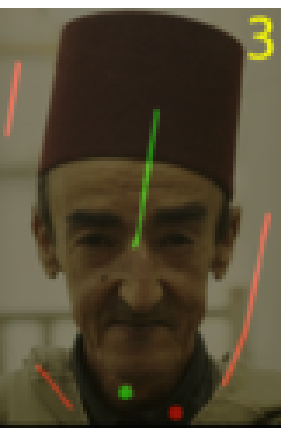}}   \hfill
		\subfloat{\includegraphics[width=0.15\textwidth, height=0.10\textwidth]{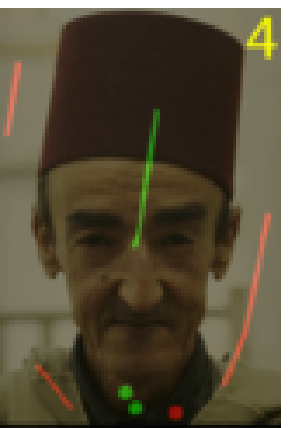}}   \hfill
	\subfloat{\includegraphics[width=0.15\textwidth, height=0.10\textwidth]{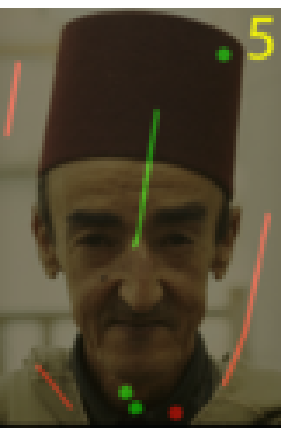}}   \hfill
	\subfloat{\includegraphics[width=0.15\textwidth, height=0.10\textwidth]{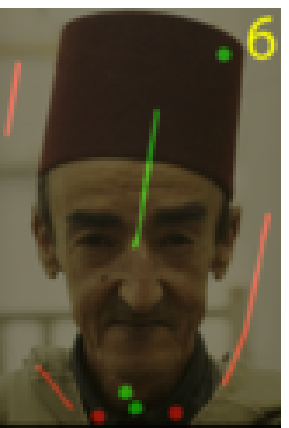}}   

	
	\subfloat{\includegraphics[width=0.15\textwidth, height=0.10\textwidth]{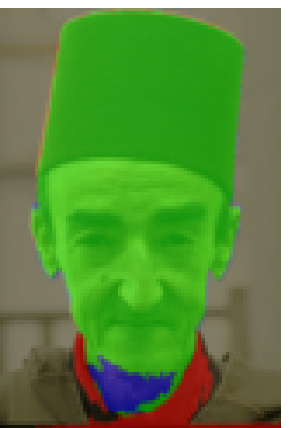}}   \hfill
		\subfloat{\includegraphics[width=0.15\textwidth, height=0.10\textwidth]{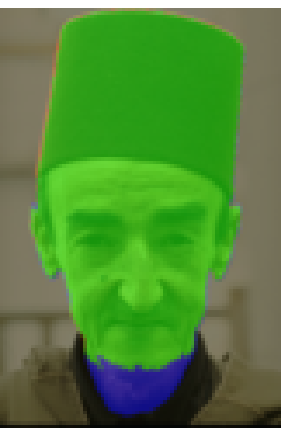}}   \hfill
	\subfloat{\includegraphics[width=0.15\textwidth, height=0.10\textwidth]{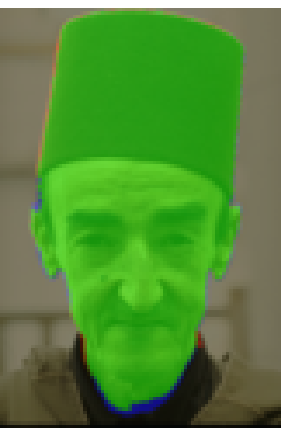}}    \hfill
		\subfloat{\includegraphics[width=0.15\textwidth, height=0.10\textwidth]{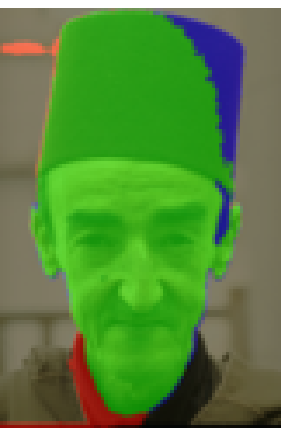}}   \hfill
		\subfloat{\includegraphics[width=0.15\textwidth, height=0.10\textwidth]{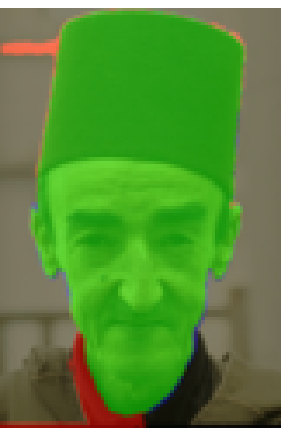}}   \hfill
		\subfloat{\includegraphics[width=0.15\textwidth, height=0.10\textwidth]{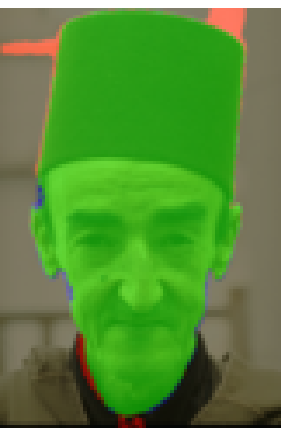}}

		\caption[Robot User Simulation]{Robot User Simulation. Green annotations indicate foreground while red annotations indicate background. Green region indicates true positive. Red region indicates false positive. Blue region indicates false negative. Best seen in zoom and color}\label{fig:Robot_User_Simulation}	
\end{figure}
\subsection{Approach Generalization}
To generalize our SL approach to other datasets, the following parameter settings are fixed, which are able to achieve the best results over Geodesic Star-Dataset:
\begin{enumerate}
  \item Number of foreground/background pivots is 21 each
  \item Number of eigenvectors is 100
\end{enumerate} 

We evaluate our approach over different datasets with the same parameter settings. Tables ~\ref{table:quality_eval_ji} and ~\ref{table:quality_eval_fs} show that SL is very competitive and superior over other segmentation approaches over all datasets. It is clear that SL is more stable as the standard deviation of the Jaccard index is superior or very close to both SP-SIG and SP-IG. Other methods, like GSC and ESC, presented as state-of-the-art in ~\cite{gulshan2010geodesic}, suffer high standard deviation.

\begin{table*}
	\scriptsize
	\centering
	\caption[Weizmann Horses Dataset Comparative Evaluation Experiment]{Quantitative Evaluation using Jaccrad Index over Weizmann Horses, BSD 100, Weizmann Single, and Weizmann Two objects datasets}
		\begin{tabular}{|l|c|c|c|c|}
					\hline 
			Method & Weizmann Horses & BSD 100 & Weizmann Single &  Weizmann Two \\
					\hline
					BJ & 0.60 $\pm$ 0.23 & 0.53 $\pm$ 0.25  & 0.66 $\pm$0.24 & 0.48 $\pm$ 0.27 \\
					\hline
					RW & 0.55 $\pm$ 0.15 & 0.49 $\pm$ 0.23  & 0.42 $\pm$ 0.26 & 0.63 $\pm$ 0.25 \\
					\hline
					PP & 0.60 $\pm$ 0.24 & 0.59 $\pm$ 0.24  & 0.68 $\pm$ 0.23 & 0.71 $\pm$ 0.25 \\
					\hline
					GSC & 0.57 $\pm$ 0.22 & 0.57 $\pm$ 0.25  & 0.69 $\pm$ 0.23 & 0.70 $\pm$ 0.25 \\
					\hline
					ESC & 0.55 $\pm$ 0.21 & 0.56 $\pm$ 0.25  & 0.67 $\pm$ 0.22 & 0.68 $\pm$ 0.25 \\
					\hline
					SP-IG & 0.51 $\pm$ 0.11 & 0.52 $\pm$ 0.15  & 0.48 $\pm$ 0.16 & 0.61 $\pm$ 0.20 \\
					\hline					
					SP-LIG & 0.48 $\pm$ 0.20 & 0.55 $\pm$ 0.23  & 0.57 $\pm$ 0.24 & 0.60 $\pm$ 0.25 \\
					\hline
					SP-SIG & 0.57 $\pm$ 0.12 & 0.59 $\pm$ 0.15  & 0.57 $\pm$ 0.18 & 0.70 $\pm$ 0.18 \\
					\hline
					\hline
					\textbf{SL} & \textbf{0.63} $\pm$ 0.15 & \textbf{0.64} $\pm$ 0.16 &  \textbf{0.72} $\pm$ 0.16 & \textbf{0.75} $\pm$ 0.17 \\
					\hline							
					\hline							
		\end{tabular}
	\label{table:quality_eval_ji}
\end{table*}

\begin{table*}
	\scriptsize
	\centering
	\caption[Weizmann Horses Dataset Comparative Evaluation Experiment]{Quantitative Evaluation using F-score over Weizmann Horses, BSD 100, Weizmann Single, and Weizmann Two objects datasets}
	\begin{tabular}{|l|c|c|c|c|}
		\hline 
		Method & Weizmann Horses & BSD 100 & Weizmann Single &  Weizmann Two \\
		\hline
		BJ & 0.72 $\pm$ 0.20 & 0.66 $\pm$ 0.22  & 0.76 $\pm$ 0.21 & 0.60 $\pm$ 0.26 \\
		\hline
		RW & 0.70 $\pm$ 0.13 & 0.63 $\pm$ 0.21  & 0.55 $\pm$ 0.26 & 0.74 $\pm$ 0.22 \\
		\hline
		PP & 0.72 $\pm$ 0.21 & 0.71 $\pm$ 0.21  & 0.78 $\pm$ 0.20 & 0.80 $\pm$ 0.22 \\
		\hline
		GSC & 0.70 $\pm$ 0.21 & 0.69 $\pm$ 0.22  & 0.79 $\pm$ 0.20 & 0.79 $\pm$ 0.22 \\
		\hline
		ESC & 0.68 $\pm$ 0.20 & 0.68 $\pm$ 0.22  & 0.78 $\pm$ 0.20 & 0.78 $\pm$ 0.22 \\
		\hline
		SP-IG & 0.67 $\pm$ 0.09 & 0.67 $\pm$ 0.13  & 0.63 $\pm$ 0.15 & 0.74 $\pm$ 0.17 \\
		\hline					
		SP-LIG & 0.63 $\pm$ 0.19 & 0.68 $\pm$ 0.19  & 0.69 $\pm$ 0.22 & 0.71 $\pm$ 0.23 \\
		\hline
		SP-SIG & 0.71 $\pm$ 0.10 & 0.73 $\pm$ 0.12  & 0.71 $\pm$ 0.16 & 0.80 $\pm$ 0.14 \\
		\hline
		\hline
		\textbf{SL} & \textbf{0.76} $\pm$ 0.11 & \textbf{0.77} $\pm$ 0.12 &  \textbf{0.82} $\pm$ 0.12 & \textbf{0.84} $\pm$ 0.13 \\
		\hline							
		\hline							
	\end{tabular}
	\label{table:quality_eval_fs}
\end{table*}

\subsection{Robot User Analysis}
\begin{figure*}
\centering
\begin{tikzpicture}
\begin{axis}[ xlabel={No. of strokes},
title=Robot User Performance Analysis,
xmin=-1,
legend pos=outer north east,
legend pos=outer north east,
 ylabel={Jaccard Index}] 

\addplot coordinates {(0,0.49)(1,0.58)(2,0.63)(3,0.66)(4,0.70)(5,0.74)(6,0.75)(7,0.76)(8,0.79)(9,0.79)(10,0.82)(11,0.82)(12,0.84)(13,0.84)(14,0.86)(15,0.86)(16,0.87)(17,0.88)(18,0.87)(19,0.88)(20,0.89)}; 

\addplot coordinates {(0,0.59)(1,0.68)(2,0.75)(3,0.78)(4,0.83)(5,0.84)(6,0.87)(7,0.89)(8,0.90)(9,0.91)(10,0.92)(11,0.93)(12,0.93)(13,0.94)(14,0.94)(15,0.95)(16,0.95)(17,0.96)(18,0.96)(19,0.96)(20,0.97)}; 

\addplot coordinates {(0,0.56)(1,0.63)(2,0.67)(3,0.72)(4,0.73)(5,0.77)(6,0.78)(7,0.80)(8,0.82)(9,0.83)(10,0.84)(11,0.86)(12,0.87)(13,0.87)(14,0.88)(15,0.89)(16,0.90)(17,0.90)(18,0.91)(19,0.91)(20,0.92)}; 

\addplot coordinates {(0,0.60)(1,0.68)(2,0.74)(3,0.76)(4,0.80)(5,0.81)(6,0.83)(7,0.85)(8,0.86)(9,0.87)(10,0.89)(11,0.89)(12,0.90)(13,0.89)(14,0.91)(15,0.92)(16,0.92)(17,0.93)(18,0.94)(19,0.94)(20,0.94)}; 

\addplot coordinates {(0,0.62)(1,0.69)(2,0.73)(3,0.77)(4,0.78)(5,0.81)(6,0.83)(7,0.83)(8,0.85)(9,0.87)(10,0.88)(11,0.88)(12,0.89)(13,0.89)(14,0.90)(15,0.92)(16,0.92)(17,0.93)(18,0.93)(19,0.93)(20,0.94)}; 

\addplot coordinates {(0,0.61)(1,0.71)(2,0.76)(3,0.81)(4,0.82)(5,0.86)(6,0.88)(7,0.89)(8,0.91)(9,0.93)(10,0.93)(11,0.94)(12,0.94)(13,0.95)(14,0.96)(15,0.96)(16,0.95)(17,0.96)(18,0.97)(19,0.97)(20,0.97)}; 

\addplot coordinates {(0,0.61)(1,0.72)(2,0.78)(3,0.81)(4,0.85)(5,0.88)(6,0.88)(7,0.89)(8,0.90)(9,0.92)(10,0.94)(11,0.95)(12,0.95)(13,0.95)(14,0.95)(15,0.96)(16,0.96)(17,0.97)(18,0.97)(19,0.97)(20,0.97)}; 


\addplot coordinates{(0,0.69)(1,0.74)(2,0.77)(3,0.81)(4,0.84)(5,0.86)(6,0.88)(7,0.89)(8,0.90)(9,0.91)(10,0.92)(11,0.93)(12,0.94)(13,0.94)(14,0.95)(15,0.95)(16,0.96)(17,0.96)(18,0.97)(19,0.97)(20,0.97) }; 

\legend{BJ,PP,SP-IG,SP-LIG,SP-SIG,GSC,ESC,SL}
\end{axis} 
\end{tikzpicture}

\begin{tikzpicture}
\begin{axis}[ xlabel={No. of strokes},
title=Robot User Stability Analysis,
xmin=-1,
legend pos=outer north east,
 ylabel={Error Margin}] 

\addplot coordinates {(0,0.26)(1,0.25)(2,0.24)(3,0.24)(4,0.22)(5,0.21)(6,0.20)(7,0.19)(8,0.18)(9,0.18)(10,0.16)(11,0.17)(12,0.15)(13,0.15)(14,0.13)(15,0.13)(16,0.13)(17,0.12)(18,0.14)(19,0.13)(20,0.11)}; 

\addplot coordinates {(0,0.25)(1,0.24)(2,0.21)(3,0.21)(4,0.17)(5,0.18)(6,0.13)(7,0.13)(8,0.11)(9,0.11)(10,0.10)(11,0.09)(12,0.10)(13,0.07)(14,0.07)(15,0.07)(16,0.07)(17,0.06)(18,0.06)(19,0.05)(20,0.04)}; 

\addplot coordinates {(0,0.16)(1,0.15)(2,0.14)(3,0.12)(4,0.12)(5,0.11)(6,0.11)(7,0.10)(8,0.09)(9,0.09)(10,0.09)(11,0.08)(12,0.08)(13,0.08)(14,0.07)(15,0.07)(16,0.06)(17,0.07)(18,0.06)(19,0.06)(20,0.06)}; 

\addplot coordinates {(0,0.22)(1,0.19)(2,0.16)(3,0.15)(4,0.14)(5,0.13)(6,0.12)(7,0.12)(8,0.11)(9,0.09)(10,0.10)(11,0.09)(12,0.09)(13,0.10)(14,0.08)(15,0.07)(16,0.06)(17,0.06)(18,0.05)(19,0.06)(20,0.06)}; 

\addplot coordinates {(0,0.17)(1,0.16)(2,0.13)(3,0.12)(4,0.11)(5,0.12)(6,0.10)(7,0.10)(8,0.09)(9,0.08)(10,0.08)(11,0.08)(12,0.09)(13,0.09)(14,0.08)(15,0.07)(16,0.06)(17,0.06)(18,0.06)(19,0.07)(20,0.06)}; 

\addplot coordinates {(0,0.26)(1,0.22)(2,0.21)(3,0.17)(4,0.18)(5,0.15)(6,0.12)(7,0.13)(8,0.11)(9,0.07)(10,0.08)(11,0.07)(12,0.08)(13,0.06)(14,0.05)(15,0.05)(16,0.09)(17,0.06)(18,0.03)(19,0.03)(20,0.04)}; 

\addplot coordinates {(0,0.25)(1,0.21)(2,0.18)(3,0.18)(4,0.13)(5,0.13)(6,0.14)(7,0.14)(8,0.13)(9,0.09)(10,0.07)(11,0.05)(12,0.05)(13,0.06)(14,0.06)(15,0.04)(16,0.07)(17,0.04)(18,0.04)(19,0.05)(20,0.03)}; 

\addplot coordinates
{(0,0.17)(1,0.15)(2,0.14)(3,0.15)(4,0.12)(5,0.11)(6,0.10)(7,0.09)(8,0.07)(9,0.07)(10,0.06)(11,0.06)(12,0.06)(13,0.06)(14,0.05)(15,0.04)(16,0.04)(17,0.03)(18,0.03)(19,0.02)(20,0.04) }; 

\legend{BJ,PP,SP-IG,SP-LIG,SP-SIG,GSC,ESC,SL}
\end{axis} 
\end{tikzpicture}

\caption{Robot User Performance Analysis over 20 strokes. The first figure shows that SL is superior to other segmentation approaches when number of strokes is low. As the number of strokes increase, segmentation task becomes easier for all approaches, yet SL achieves steady competitive results. The second figure studies the stability of segmentation accuracy as number of strokes increase. SL shows better reliability by fulfilling lower standard deviation.}
  \label{fig:robot_analysis}
\end{figure*}
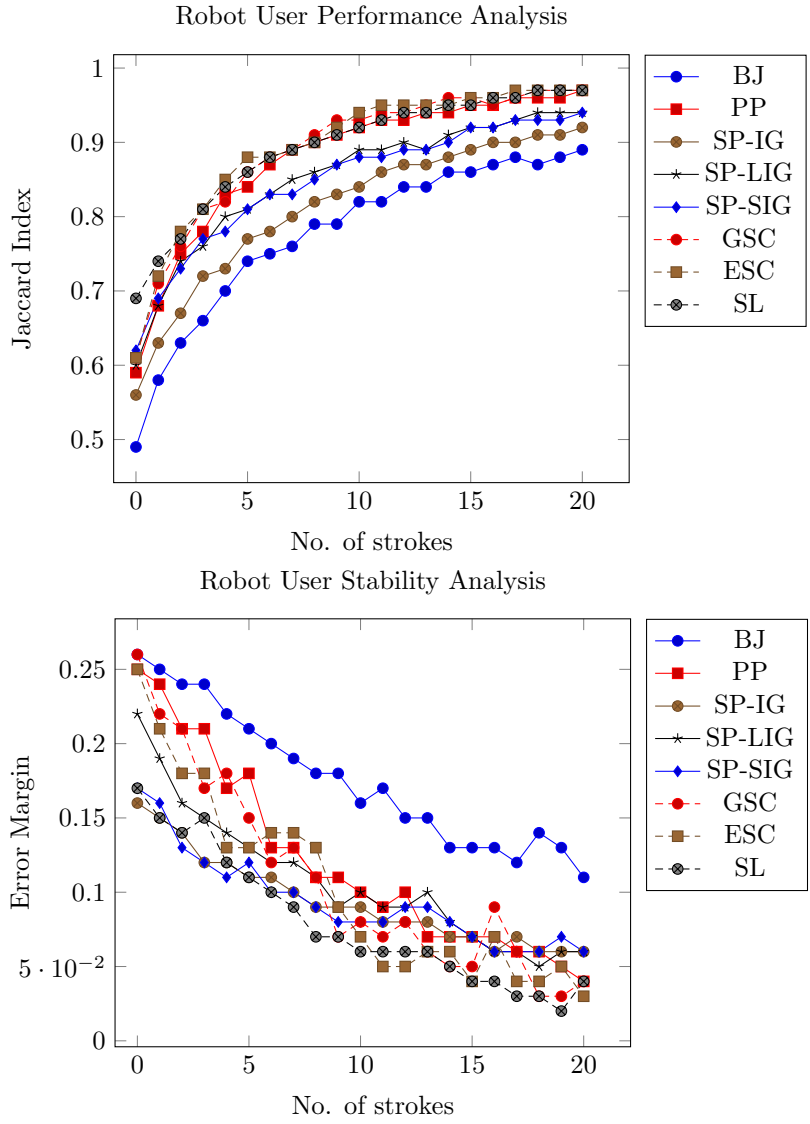

Jaccard index and standard deviation metrics are used to evaluate the performance of SL against other segmentation methods. Figure ~\ref{fig:Robot_User_Simulation} shows how SL adapts to the robot user annotations and updates the segmentation result accordingly. Figure \ref{fig:robot_analysis} shows a quantitative evaluation. SL outperforms other approaches when the number of strokes is low and is very competitive when the number of strokes increases. 


To study SL stability, we use the standard deviation measure. Figure  \ref{fig:robot_analysis} shows a comparison between SL and other segmentation methods. This figure shows that the standard deviation measure of SL is always lower than that of other approaches. This finding shows SL stability and reliability. Table \ref{table:robot_user_avg_effort} shows the interaction effort required to reach an accuracy band between [85 98] in terms of brush strokes count.

\begin{table}
\centering
\caption[Average User Effort Analysis]{User Average Effort Analysis. We compare the avg. No. of user strokes needed for every segmentation approach to reach a certain quality band [$Alow=0.85$, $Ahigh=0.98$].}

\begin{tabular}{|c|c|c|c|c|c|c||c|}
\hline
BJ & PP & GSC & ESC & SP-IG & SP-LIG & SP-SIG & SL \\ \hline
19.68 & 11.57 &  10.54 & \textbf{10.24} & 17.83 &15.19 & 15.74 & 10.51  \\ \hline
\end{tabular}
\label{table:robot_user_avg_effort}
\end{table}

  